\documentclass{article}

    \PassOptionsToPackage{numbers, compress}{natbib}

\usepackage[preprint]{neurips_2025}

\usepackage[utf8]{inputenc} 
\usepackage[T1]{fontenc}    
\usepackage{hyperref}       
\usepackage{url}            
\usepackage{booktabs}       
\usepackage{amsfonts}       
\usepackage{nicefrac}       
\usepackage{microtype}      
\usepackage{amsmath,comment}
\usepackage{booktabs}
\usepackage{multirow}
\usepackage{graphicx}
\usepackage{enumitem}
\usepackage{listings}
\usepackage{wrapfig}
\usepackage{tablefootnote}
\usepackage{pifont}

\usepackage{caption}
\usepackage{algorithm}
\usepackage{algpseudocode} 

\usepackage[table,xcdraw]{xcolor}

\definecolor{toponeRed}{HTML}{FF9999}  
\definecolor{toptwoRed}{HTML}{FFCCCC}  
\definecolor{mylightgray}{HTML}{E8E8E8} 

\newcommand{\cmark}{\ding{51}}   
\newcommand{\xmark}{\ding{55}}   

\definecolor{myForestGreen}{RGB}{34,139,34}

\title{The \emph{Avengers}: A Simple Recipe for Uniting Smaller Language Models to Challenge Proprietary Giants}

\author{
    \textbf{Yiqun~Zhang}$^{1, 2\clubsuit\ddagger}$ \quad
    \textbf{Hao~Li}$^{2,3\clubsuit\ddagger}$ \quad
    \textbf{Chenxu~Wang}$^{2,4 \ddagger}$ \quad
    \textbf{Linyao~Chen}$^{2,5 \ddagger}$ \\\
    \textbf{Qiaosheng~Zhang}$^2$ \quad
    \textbf{Peng~Ye}$^2$ \quad 
    \textbf{Shi~Feng}$^{1\dag}$ \quad 
    \textbf{Daling~Wang}$^{1}$ \quad 
    \textbf{Zhen~Wang}$^3$ \\
    \textbf{Xinrun~Wang}$^6$ \quad
    \textbf{Jia~Xu}$^2$ \quad
    \textbf{Lei~Bai}$^2$ \quad 
    \textbf{Wanli~Ouyang}$^2$ \quad
    \textbf{Shuyue~Hu}$^{2\dag}$ \\
    $^1$ Northeastern University \quad $^2$
     Shanghai Artificial Intelligence Laboratory \\ $^3$ Northwestern Polytechnical University \quad
    $^4$ Beijing Institute of Technology \\ $^5$ The University of Tokyo 
    \quad $^6$ Singapore Management University \\
    \texttt{yiqunzhang@stumail.neu.edu.cn} \quad   \texttt{li.hao@mail.nwpu.edu.cn} \\   
    \texttt{fengshi@cse.neu.edu.cn} \quad
    \texttt{hushuyue@pjlab.org.cn}
}

\begin{document}
\maketitle
\vspace{-2em}
\begin{abstract}
\vspace{-1em}
Proprietary giants are increasingly dominating the race for ever-larger language models. Can open-source, smaller models remain competitive across a broad range of tasks? 
In this paper, we present the \emph{Avengers}---a simple recipe that  leverages the collective intelligence of these smaller models. The \emph{Avengers} builds upon four lightweight operations: (i) \textit{embedding}: encode queries using a text embedding model;
(ii) \textit{clustering}: group queries based on their semantic similarity;
(iii) \textit{scoring}: scores each model's performance within each cluster;
and (iv) \textit{voting}: improve outputs via repeated sampling and voting.
At inference time, each query is embedded and assigned to its nearest cluster. The top-performing model(s) within that cluster are selected to generate the response with repeated sampling.
Remarkably, \textbf{with 10 open-source models ($\sim $7B parameters each), the \emph{Avengers} surpasses GPT-4o, 4.1, and 4.5 in average performance across 15 diverse datasets spanning mathematics, coding, logical reasoning, general knowledge, and affective tasks.} In particular, it surpasses GPT-4.1 on mathematics tasks by 18.21\% and on code tasks by 7.46\%. 
Furthermore, the \emph{Avengers} delivers superior out-of-distribution generalization, and remains robust across various embedding models, clustering algorithms, ensemble strategies, and values of its sole parameter---the number of clusters.  
We have open-sourced the code on GitHub: 
\url{https://github.com/ZhangYiqun018/Avengers}
\end{abstract}
\vspace{-2.0em}
\begin{center}
\includegraphics[width=1\textwidth]{figures/performance_4model_typicalbench.pdf}
\captionof{figure}{Comparison of proprietary models with our method (the \emph{Avengers}) across six representative benchmarks. Full results on all 15 benchmarks are reported in Table~\ref{tab:main-result}.}
\label{fig:main-result-figure}
\vspace{-1em}
\end{center}

\let\thefootnoteorig\thefootnote
\def\thefootnote{\arabic{footnote}} 
\def\thefootnote{\dag}\footnotetext{Corresponding author.}
\def\thefootnote{$\clubsuit$}\footnotetext{Equal contribution.}
\def\thefootnote{$\ddagger$}\footnotetext{Work done during the author’s internship at Shanghai Artificial Intelligence Laboratory.}
\let\thefootnote\thefootnoteorig

\section{Introduction}
The current landscape of language models (LMs) is dominated by few proprietary giants. The race to build ever-larger models is far from over; however, it is becoming a closed competition, accessible only to organizations with massive resources. This trend risks marginalizing the broader research community and concentrating innovation within a privileged few. 
In this context, the open-source community---typically working with \textit{smaller} LMs---must ask whether it is able to remain competitive. 
Although recent studies have shown that individual smaller models can excel in specific domains~\citep{liu2025fin,hui2024qwen2,zhang2024ultramedical,cui2025process}, the more pressing challenge is: can it remain competitive not only on narrowly defined tasks, but also in achieving comparable or even superior \emph{overall} performance across a \emph{broad} range of tasks?

A positive answer would have important implications. First, since open-source, smaller models are easily accessible and significantly more amenable, a positive answer would represent a meaningful step toward democratizing AI development~\citep{vryn2025community}---enabling a broader community of researchers and practitioners to engage in and contribute to cutting-edge LM research.
Second, the open-source ecosystem has already invested significant effort and computational resources in model development.\footnote{For example, as of May 2025, there are over 1,318 models fine-tuned on LLaMA-3.1-8B-Instruct and 489 on Qwen-2.5-7B-Instruct, despite both being released less than a year ago.} Demonstrating strong performance from these models would justify the reuse of existing efforts, supporting more sustainable AI research~\citep{schwartz2020green}.
Third, a positive answer would pave a new path for advancing LM research,  providing another viable alternative to the ever increasing of model size.

We posit that the key to addressing the challenge lies in \emph{collective intelligence}---while a single smaller model may not match the generality of larger ones, the collective performance of a team of multiple smaller models may prove competitive. Nevertheless, operationalizing collective intelligence is non-trivial and involves several important design considerations. Since it is infeasible to incorporate all available models in the community, which subset of models should be selected? How should we orchestrate this subset of models to maximize their collective performance? For a notable example, for each incoming query, should we rely on a router-based approach~\citep{jiang2023llm,chen2024routerdc,zhuang2024embedllm,zhang2025capability} that selects the most capable model, or should we adopt a mixture-based strategy (e.g., Mixture-of-Agents~\citep{wang2024mixture,li2024smoa,li2025rethinking,chen2025symbolicmixtureofexpertsadaptiveskillbased}) where multiple models respond in parallel and their outputs are aggregated into a final answer?
Moreover, given the rapid evolution of the LM landscape, with new tasks and new models continuously emerging, how can such systems adapt to these ongoing changes?



\begin{figure}[t]
    \centering
    \includegraphics[width=1\linewidth]{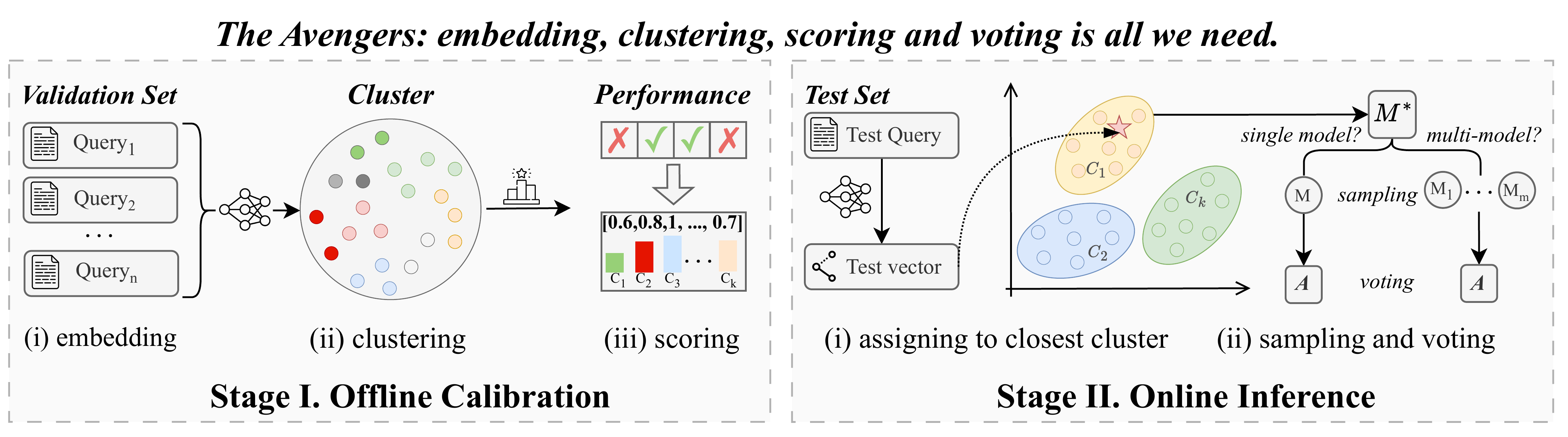}
    \caption{The \emph{Avengers}, a simple recipe for \emph{collective intelligence} of smaller language models.}
    \label{fig:main-figure}
      \vspace{-5pt}
\end{figure}

To this end, we introduce the \emph{Avengers}---a simple recipe that effectively leverages the collective intelligence of smaller LMs.
Given a set of models and a dataset a dataset (potentially spanning multiple tasks and benchmarks), the \emph{Avengers} operates as follows. First, it analyzes the tasks by embedding queries from the validation set and clustering them based on their semantic representations. Next, to identify each model's strengths and weaknesses, it evaluates each model on the validation set and constructs a cluster-wise capability profile---a simple, numeric vector representing model performance across clusters. At inference time, for each query in the test set, the \emph{Avengers} computes its embedding and assigns it to the nearest cluster. Based on the capability profiles, the model(s) with the highest performance in that cluster are then selected to handle the query. The final output is generated using a repeated sampling and voting strategy over the selected model(s). An overview of the \emph{Avengers} is presented in Figure~\ref{fig:main-figure}.


The \emph{Avengers} is straightforward to implement, requiring no neural network training. It is highly reproducible, avoiding reliance on hand-crafted prompts or architecture-specific model choices. It is plug-and-play, allowing seamless integration of newly available models by incrementally evaluating them on the validation set and computing their capability profiles. Moreover, it is well-suited for cold-start,  assuming no human expertise or prior knowledge for model set composition; given a large candidate model pool, the construction of capability profiles enables automatic selection of high-performing models while preserving diversity. Conceptually, the \emph{Avengers} can be viewed as a routing-based approach. However, as detailed in the Related Work section and summarized in the qualitative comparison in Table 1, it differs significantly from prior work in this category---both methodologically, through its clustering-based routing strategy, and empirically, in the notably strong performance it achieves.

Our experiments show that, the \emph{Avengers}, using 10 models ($\sim$7B each), surpasses the average performance of GPT-4.1\footnote{\href{https://openai.com/index/gpt-4-1/}{GPT-4.1}: OpenAI's flagship model released on April 14, 2025.} across 15 datasets spanning five domains (mathematics, code, logic, knowledge, and affective tasks). Notably, the \emph{Avengers} outperforms GPT-4.1 on nine of these datasets;  it outperforms GPT-4.1 by $37.04\%$ on AIME, by $31.86\%$ on MBPP, and by $15.41\%$ on KORBench.
When isolating the effect of the routing mechanism, the \emph{Avengers}' clustering-based router still outperforms state-of-the-art (SOTA) performance-oriented routers (RouterDC~\citep{chen2024routerdc}, embedLLM~\citep{zhuang2024embedllm}, and MODEL-SAT~\citep{zhang2025capability}), even though all those routers require neural network training (with some even involving fine-tuning LMs) and must be retrained to accommodate new models or tasks. 
On out-of-distribution tasks (evaluated on five additional benchmarks), the superior performance of the \emph{Avengers}' clustering-based router is even more noticeable; it outperforms RouterDC by 10.31\%, embedLLM by 8.14\%, and MODEL-SAT by 8.56\%, showcasing its strong generalization capability.


Additional experiments demonstrate that the \emph{Avengers} is compatible with a variety of embedding models, clustering algorithms, and ensemble strategies for aggregating outputs from selected models. It maintains robust performance across different clustering methods (K-Means~\citep{lloyd1982least, macqueen1967some}, Hierarchical Clustering~\citep{ward1963hierarchical}, GMM~\citep{dempster1977maximum}, Spectral Clustering~\citep{von2007tutorial} and BIRCH ~\citep{zhang1996birch}). Furthermore, its performance can be further enhanced by employing stronger embedding models and ensemble strategies that explicitly leverage multiple models, such as the multi-model variant~\citep{chen2025trulyneedsamplesmultillm} of the repeated-sampling-and-voting approach.
Furthermore, we show that the \emph{Avengers} can automatically construct an effective model set given a predefined set size. Given a candidate pool of 22 models, it can match the performance of GPT-4.1 with as few as three selected models, yet its performance continues to improve as the number of models increases, particularly in Code, Knowledge, and Affective tasks (Figure~\ref{fig:modelselection_mp}). 
In addition, regarding data efficiency, we show that the \emph{Avengers} continues to outperform GPT-4.1 even when only 30\% of the data is used for validation, leaving the remaining 70\% for testing (Figure~\ref{fig:testsize}). Finally, we show that the \emph{Avengers} is robust to variations in its sole hyperparameter, i.e. the number of clusters $k$. Building on this observation, we offer a hypothesis to explain why clustering-based routing can outperform typical supervised routing methods, despite requiring no training (Appendix~\ref{app:insight}, Figure~\ref{fig:k_impact}).

An exploratory analysis (Appendix~\ref{app:model-usage}, Figure~\ref{fig:model-usage-distribution})) about the working mechanism of the \emph{Avengers} reveals that it can automatically select models that strike a favorable balance between strong individual performance and diversity, which lays the foundation for accurate query-to-model routing.  From a candidate pool, it can select models, such as Fin-R1 and LLaMA-3.1-8B-UltraMedical, which are of rather low visibility in the community. However, these models prove crucial to overall performance. Notably,  $\sim$50\% of queries from the ARC Challenge are routed to fin-R1 (originally developed for financial tasks), and  $\sim$40\% of queries from MedQA are routed to LLaMA-3.1-8B-UltraMedical (developed by a small group of university students). These results not only demonstrate the effectiveness of the \emph{Avengers}, but also more importantly, highlight the untapped potential of open-source contributions---advancing the vision of a more democratic and inclusive AI ecosystem.

    





\section{Related Work}

The harnessing of collective intelligence from multiple models constitutes one of the frontiers of AI and ML research, and has recently attracted much interest~\citep{lu2024merge,guo2024large,zhang2025if,subramaniam2025multiagent,wan2025rema,zheng2025decouple}. Existing approaches in this area generally fall into three paradigms: router-based, mixture-based, and merging-based methods.
This work is most closely aligned with the router-based paradigm. Due to space constraints, we focus our discussion on this line of research and defer coverage of the other two paradigms to the Appendix.
 
The main goal of most router-based methods is to enhance the overall performance of a set of smaller models through query routing; a neural network-based router is often trained to select, for each incoming query, the model most capable of handling it~\citep{chenfrugalgpt,shnitzerlarge,ong2024routellm,feng2025graphrouter,shen2024learning,huang2025routereval}.
LLM-Blender~\citep{jiang2023llm} utilizes pairwise comparisons to select the top-\( k \) models for each query and then fuses their outputs to enhance performance. ZOOTER~\citep{lu2024routing} proposes reward-guided query routing, employing tag-based label enhancement to improve training stability.
More recently, RouterDC~\citep{chen2024routerdc}, which employs dual contrastive learning to enhance routing accuracy, and EmbedLLM~\citep{zhuang2024embedllm} leverages learned compact model embeddings along with query embeddings to predict routing correctness. Additionally, Model-SAT~\citep{zhang2025capability} generates capability instructions from model aptitude outcomes and employs text-aligned embeddings to guide a lightweight LLM in selecting optimal candidate models. 

In parallel with the above performance-oriented research, a few recent studies have also explored routing strategies that account for the trade-off between performance and computational cost. 
Routellm~\citep{ong2024routellm} trains a binary classifier using preference data to dynamically route queries to either a stronger or a weaker LLM during inference.
GraphRouter~\citep{feng2025graphrouter} constructs a heterogeneous graph comprising task, query, and LLM nodes, and predicts the performance-cost score via an edge prediction mechanism.
RouterBench~\citep{hu2024routerbench} introduces a suite of benchmarks along with several routing strategies. One such strategy routes each query to an LLM based on the average cost-penalized performance of the query's k nearest neighbors.  
Similarly, \citep{jitkrittum2025universal} introduces a cluster-based approach that groups queries using a large unlabeled dataset with K-means; each query is assigned to a cluster, and an LLM is selected based on cost-adjusted per-cluster error scores computed from another labeled dataset.

The \emph{Avengers} falls within the router-based paradigm but departs significantly from prior work in two important ways. First, unlike most existing methods that rely on training neural routers, which must be re-trained to accommodate new tasks or models---a significant limitation given the rapid pace of LLM development. In contrast, our approach is entirely \textit{training-free}, yet remains highly adaptable and competitive, offering a lightweight, reproducible alternative without sacrificing performance. 
Second, while training-free routing (e.g., $K$-NN or $K$-means routing) has been explored in cost-performance tradeoff settings,  they typically operate on model pools with large size disparities (often including models far larger than those used in our work), yet still \textit{fail} to demonstrate competitive performance. To the best of our knowledge, this is the first work to show that a training-free, router-based framework---when integrated with automatic complementary model selection and voting---can elevate small open-source models to match the \emph{overall} performance of a proprietary, flagship LLM.










\section{The \emph{Avengers}}
\label{method}
In this section, we introduce the \emph{Avengers}, a simple yet effective framework for harnessing the collective intelligence of multiple smaller language models.
The \emph{Avengers} is based on the observation that the open-source ecosystem offers a wide range of small models, with growing evidence that these models exhibit substantial diversity. This diversity arises in two key ways: (i) some models are domain-specific experts, fine-tuned for particular tasks such as mathematical reasoning; and (ii) others, while general-purpose, differ due to variations in training data, architectures, and design choices made by different stakeholders. 

This diversity provides a strong foundation for collective intelligence. However, operationalizing it requires careful consideration of several key design decisions. A core principle underlying the \emph{Avengers} is the idea of ``\textbf{horses for courses}''—selecting the most suitable model(s) for each incoming query and routing the query to that model. We prioritize a routing strategy over a mixture strategy,\footnote{Mixture-based methods often demonstrate greater performance gains over single models, compared to router-based approaches.} because unlike larger models, smaller models tend to be more sensitive to prompt variations and often struggle with instruction following. Prior work has shown that mixture-based methods, while effective with larger models, can lead to degraded performance when applied to smaller ones.

The question then becomes: how do we identify the right ``horses'' for the right ``courses''? 
This requires understanding both (i) what models (``horses'') are good at, and (ii) what types of queries (``courses'') are being handled.
Moreover, it also necessitates effective matching---routing each query to the model best suited for its type. 
Consider a set $M$ of models and a set $\mathcal{D}$ of query-answer sample pairs, which can be split into a validation set $\mathcal{D}_{\text{val}}$ and a test set $\mathcal{D}_{\text{test}}$. 
The \emph{Avengers} address this question in two stages: (i) \textbf{offline calibration}, where it builds an understanding of both query types and model capabilities based on $\mathcal{D}_{\text{val}}$, and (ii) \textbf{online inference}, where it dynamically routes each query from $\mathcal{D}_{\text{test}}$ to the model it deems most capable and enhances the final output.

\textbf{Offline Calibration }
During this stage, the \emph{Avengers} constructs a structured understanding of the query type via embedding and clustering, and characterizes model capabilities through cluster-wise performance scores. 
First, to characterize query types, each query $d \in \mathcal{D}_{\text{val}}$ is encoded using a text \textbf{embedding} model, producing a semantic vector representation. Then, these vectors are clustered into $k$ groups with the use of a \textbf{clustering} method (e.g., $K$-means), which results in a set $\mathcal{C}$ of $k$ distinct clusters. Each cluster $c\in \mathcal{C}$ represents a semantically coherent group of queries, i.e., a query type. 
Next, to characterize model capabilities, each model $m\in \mathcal{M}$ is evaluated on $\mathcal{D}_{\text{val}}$, and its performance score is recorded within each cluster. This yields a cluster-wise \textbf{capability profile} for each model, which can be represented as a vector $p=[p_1,\cdots, p_k]$, where $p_i$ denotes the model’s performance score on cluster $c_i \in \mathcal{C}$. These profiles inform routing decisions at the inference time.

\textbf{Online Inference } At the inference time, the \emph{Avengers} no longer performs clustering or profiling. 
For each query in the test set $\mathcal{D}_{\text{test}}$, the \emph{Avengers} first computes its \textbf{embedding} using the same text embedding model used during offline calibration. The query is then \textbf{routed} to the nearest cluster $c_{\ast} \in \mathcal{C}$, based on distance in the embedding space.
Given the nearest cluster $c_\ast$, the \emph{Avengers} consult the capability profiles and select the top-$n$ model(s) with the highest performance score $p_{\ast}$ within that cluster. Notably, the selected model(s) is not necessarily the one with the best overall performance on $\mathcal{D}_{\text{val}}$ or $\mathcal{D}_{\text{test}}$. Rather, it is selected for its strength in the specific query type (or cluster). Once the top-$n$ model(s) is selected, the \emph{Avengers} generates responses using \textbf{repeated sampling}, followed by majority \textbf{voting} to determine the final output. That is, when $n=1$, it adopts the Self-Consistency (SC)~\citep{wang2022self}. When $n>1$, it adopts the Model-Switch~\citep{chen2025trulyneedsamplesmultillm}, a multi-model, sample-efficient extension of SC.

\textbf{
Automatic Model Set Construction }
It is worth mentioning that the composition of the model set $\mathcal{M}$ can significantly influence collective performance. Intuitively, strong overall performance is unlikely to emerge from a group of models that underperform across all tasks. While prior approaches often rely on human expertise or domain knowledge to curate such model sets, the \emph{Avengers} automates the selection of models with complementary strengths.
Given a predefined deployment budget of $|\mathcal{M}|$ models and a larger candidate pool $\mathcal{M}'$, the \emph{Avengers} enables automatic model selection from $\mathcal{M}'$. During the offline calibration stage, for each candidate model $m' \in \mathcal{M}'$, the \emph{Avengers} computes the cluster-wise capability profile and calculates an overall score across clusters: $s(m') = \sum_{c \in \mathcal{C}} 1/r^c_{m'}$, where $r^c_{m'}$ denotes the rank of model $m'$ within cluster $c$, determined by its cluster-wise performance.
A higher score 
$s(m')$ indicates that the model either excels in specific clusters (yielding high ranks) or maintains strong performance across multiple query types.
By selecting the top 
$|\mathcal{M}|$ models with the highest overall scores, the \emph{Avengers} naturally constructs a complementary model set $\mathcal{M}$ that balances specialization and diversity across different query types. 

\textbf{New Datasets and Newly Available Models }
The fast-paced evolution of LM research regularly introduces new datasets, tasks, and models. While prior routing-based methods often require retraining neural networks to accommodate such changes, the \emph{Avengers} is designed to be highly adaptable. To incorporate a new, out-of-distribution dataset $\mathcal{D}'$, the \emph{Avengers} re-executes the offline calibration stage. This requires only \emph{incremental} re-evaluation of existing models $\mathcal{M}$ on the validation set $\mathcal{D}'_{\text{val}}$, followed by re-clustering the queries from $\mathcal{D}'_{\text{val}}\cup \mathcal{D}_{\text{val}}$ based on updated query embeddings. Incorporating a newly available model $m^\dagger$ is even more lightweight. The \emph{Avengers} simply \emph{incrementally} evaluates the new model $m^\dagger$ on the existing validation set $\mathcal{D}_{\text{val}}$ and then computes its cluster-level performance profile, even without re-clustering. 

\section{Experiments}
\label{sec: experiments}

\subsection{Experimental Setup}

\paragraph{Datasets} We primarily consider 15 datasets covering five categories: \textbf{Mathematics} (AIME, Math500~\citep{lightman2023lets}, and LiveMathBench~\citep{liu2024livemathbench}), \textbf{Code} (MBPP~\citep{austin2021mbpp}  and HumanEval~\citep{chen2021evaluating}), \textbf{Logic} (KORBench~\citep{ma2024korbenchbenchmarkinglanguagemodels}, Knights and Knaves~\citep{xie2024memorization}, and BBH~\citep{suzgun2022challenging}), \textbf{Knowledge} (ARC Challenge~\citep{clark2018think}, MMLUPro~\citep{wang2024mmlupro}, GPQA~\citep{rein2024gpqa}, FinQA~\citep{chen2021finqa}, and MedQA~\citep{jin2021disease}), and \textbf{Affective} (EmoryNLP~\citep{byrkjeland2018ternary} and MELD~\citep{poria2019meldmultimodalmultipartydataset}). Additionally, to evaluate the generalization capabilities of the \emph{Avengers} on out-of-distribution (OOD) tasks, we further include one extra dataset for each category: MathBench~\citep{liu2024mathbench}, StudentEval~\citep{babe2023studenteval}, Winogrande~\citep{sakaguchi2021winogrande}, BRAINTEASER~\citep{rahimi2024nimz}, and DailyDialog~\citep{li2017dailydialog}.
Note that the additional five datasets are used exclusively for evaluating the OOD generalization.
Detailed information about these datasets is provided in Appendix~\ref{app:dataset} and Table~\ref{tab:benchmarks}.

\paragraph{Implementation of the \emph{Avengers}}
We uniformly employ \emph{gte-qwen2-7B-instruct} as the embedding model and utilize the classic $K$-Means with the number of clusters set to $K=64$. Each dataset is randomly split into 70\% for fitting cluster centroids and the remaining 30\% for performance evaluation. Initially, we consider a pool consisting of 22 open-source candidate LLMs ($\sim 7B$), from which $m=10$ models are automatically selected. 
During inference, we adopt the Self-Consistency (SC) strategy by default, setting the number of sampling rounds to 10.  
All experiments are repeated five times using five random seeds (42, 999, 2024, 2025, and 3407), and average results are reported.
Detailed model information and deployment specifics can be found in the Appendix~\ref{sec:model}.


\paragraph{Baselines}
We select baseline methods from two distinct technical directions (mixture- and router- based) to comprehensively evaluate the effectiveness of the \emph{Avengers}.
Additionally, to provide intuitive performance benchmarks, we include three proprietary models, \textbf{gpt-4o-2024-08-06} (GPT-4o), \textbf{gpt-4.5-preview} (GPT-4.5-preview) and \textbf{gpt-4.1-2025-04-14} (GPT-4.1).
Specifically, we consider the following baseline methods:
\begin{itemize}[leftmargin=*]
\setlength\itemsep{0em}
    \item \textbf{Random Router}: Randomly selects one model from candidate set $S$ for each query.
    \item \textbf{LLM Router}: A prompt-based router that uses an LLM (Qwen2.5-7B-Instruct) to select models based on natural-language profiles summarizing candidate model performance.
    \item \textbf{RouterDC}~\citep{chen2024routerdc}: Utilizes dual contrastive learning to train embeddings for queries and models. The contrastive objectives ensure query embeddings are close to suitable models and semantically similar queries cluster together.
    \item \textbf{EmbedLLM}~\citep{zhuang2024embedllm}: An embedding-based router trained via binary cross-entropy loss to predict model-query compatibility. 
    \item \textbf{MODEL‑SAT}~\citep{zhang2025capability}: Employs capability instruction tuning, translating candidate models’ performance into textual capability descriptions, which are embedded and passed to a trainable LLM to predict model suitability for each query.
    \item \textbf{Mixture of Agents (MoA)}~\citep{wang2024mixture}: A hierarchical mixture approach where multiple layers of LLMs iteratively refine or aggregate candidate answers, culminating in a final aggregation step.
    \item \textbf{Symbolic‑MoE (SMoE)}~\citep{chen2025symbolicmixtureofexpertsadaptiveskillbased}: A mixture method using skill keyword matching. Candidate models are selected based on query-derived skill keywords, and their outputs are aggregated via a single-layer MoA.
\end{itemize}

The methodological differences between these baseline methods and the \emph{Avengers} are summarized in Table~\ref{tab:methodology-comparison}. More detailed implementation specifics can be found in Appendix~\ref{baseline}.

\begin{table}[t]
    \centering
\caption{Methodological comparison between baseline methods and the \emph{Avengers}.
    \textbf{Trainable NN}: additional neural components requiring training.
    \textbf{Arch.}: effort in architecture design.
    \textbf{Prompt Eng.}: effort in prompt design.
    \textbf{New Task}: how to adapt to new tasks.
    \textbf{New Model}: how to adapt to new models.
    \textbf{Model Scale}: can scale with the number of SLMs.
    \textbf{Auto Init. Sel.}: can automatically select the initial SLM subset.
}
  \renewcommand{\arraystretch}{1.2}
  \setlength{\tabcolsep}{3pt}
    \resizebox{\columnwidth}{!}{
\begin{tabular}{@{}l c c c c c c c@{}}
    \toprule
    \textbf{Method} & 
    \textbf{Trainable NN\textsuperscript{†}} & 
    \textbf{Arch.} & 
    \textbf{Prompt} & 
    \textbf{New Task \textsuperscript{‡}} & 
    \textbf{New Model\textsuperscript{‡}} &
    \textbf{Model Scale\textsuperscript{§}} & 
    \textbf{Auto Init. Sel.} \\
    \midrule
    RouterDC~\citep{chen2024routerdc}     & Emb+MLP       & Free           & Free                 & \multicolumn{2}{c}{LS+Retrain NN} & \cmark & \xmark \\
    EmbedLLM~\citep{zhuang2024embedllm}   & MLP           & Free           & Free                 & \multicolumn{2}{c}{LS+Retrain NN} & \cmark & \xmark \\
    MODEL-SAT~\citep{zhang2025capability} & Emb+MLP+SLM   & Free           & Hand-Craft           & \multicolumn{2}{c}{LS+Retrain NN} & \cmark & \xmark \\
    \midrule
    MoA~\citep{wang2024mixture}           & Free          & Hand-Craft     & Hand-Craft           & \multicolumn{2}{c}{Free}          & \xmark & \xmark \\
    Symb.-MoE~\citep{chen2025symbolicmixtureofexpertsadaptiveskillbased} & Free & Free & Hand-Craft & \multicolumn{2}{c}{LS+Regen Profile} & \cmark & \xmark \\
    \midrule
    \textbf{Avenger (ours)}              & Free          & Free           & Free                 & LS+Recluster & LS & \cmark & \cmark \\
    \bottomrule
\end{tabular}
    }
    \label{tab:methodology-comparison}
    \vspace{0.2ex} 
    \begin{minipage}{\columnwidth} 
    \footnotesize 
    \textsuperscript{†} Emb = trainable embedding model; MLP = small feed-forward head. 
    \textsuperscript{‡} LS = labeled samples. \\
    \textsuperscript{§} Mixture of Agents exceeds the maximum context window for SLMs (e.g., 8192 tokens for Gemma-2-9B) as the number of SLMs increases.  
    \end{minipage}

\end{table}

\paragraph{Enhancement to the Baselines}
Notably, to better align the baseline methods with our experimental scenarios and facilitate comprehensive comparisons, we introduce targeted enhancements for certain baselines. All baseline methods utilize \textbf{the same set of 10 models} automatically selected by the \emph{Avengers} framework.
Specifically, for the Mixture-based method MoA, we additionally evaluate an oracle variant—MoA (Oracle)—in which we manually \textbf{select the top-3 performing models} per task instead of using all 10 models. This approach addresses the performance degradation arising from excessively long context windows in the original MoA setup. For the trainable router-based methods (RouterDC, EmbedLLM, and MODEL-SAT), we do not employ a fixed training step count as a stopping criterion; instead, we report each method’s performance at its \textbf{peak on the test set}. Additionally, all SLMs and router-based results reported in Table~\ref{tab:main-result} and~\ref{tab:ood-result} \textbf{incorporate the same Self-Consistency} (SC) strategy used by the \emph{Avengers}, with 10 samples per query, ensuring fairness and robustness across comparisons.
Note that all the aforementioned enhancements are not part of the original designs of the baseline methods. However, we empirically find that they significantly improve baseline performance.


\subsection{The \emph{Avengers} Achieves State-Of-The-Art Performance}

Table~\ref{tab:main-result} presents a detailed performance comparison between the \emph{Avengers} and three groups of baselines across 15 datasets in five task categories. 
First, the \emph{Avengers} \textbf{achieves an average score of 70.54, surpassing GPT-4.1’s 69.20}, demonstrating that dynamically selecting among smaller models yields performance comparable to large-scale single models. Specifically, the \emph{Avengers} outperforms GPT-4.1 on 10 of the 15 datasets, achieving an average improvement of 18.21\% on mathematical tasks, with notable improvements of 31.86\% on MBPP and 15.41\% on KORBench. However, GPT-4.1 retains its advantage on specialized knowledge-intensive tasks (e.g., GPQA, MedQA) due to its larger parameter size and broader knowledge base. Nevertheless, the \emph{Avengers} exceeds the max expert by 6.89\% on MMLUPro (61.94 to 66.21), nearly matching GPT-4.1, underscoring the complementary potential of smaller models.

Second, mixture-based methods (MoA, SMoE, and MoA (Oracle)) struggle under our setting. Under identical candidate model pools, router-based methods generally outperform mixture-based methods. This is primarily due to limited model context windows and weaker instruction-following abilities of smaller models, resulting in lower performance ($<$ 61) compared to the \emph{Avengers}.

Third, even when we enhance router-based methods by selecting their peak on test sets, the \emph{Avengers} still \textbf{achieves the best average performance }without involving additional training parameters. 
Moreover, traditional router-based methods often struggle in Out-of-Distribution (OOD) settings.
To compare their generalization ability, we introduce a new dataset for each of the five task categories to evaluate performance. As shown in Table~\ref{tab:ood-result}, the \emph{Avengers} \textbf{exhibits the most robust generalization }among all compared methods, achieving an average score of 74.42. This performance surpasses the best router-based baseline by EmbedLLM (+8.14\%).

It should be noted that the above baselines results have been enhanced with techniques mentioned in the last section. We choose to report results with these enhancements, rather than strictly adhering to the original implementations, for two reasons:
(i) to isolate the impact of the routing mechanism itself, enabling a fairer comparison between our clustering-based approach and other routing methods; and
(ii) to contribute constructively to the field by demonstrating that some of the performance gains achieved by the {Avengers} can, in principle, be reproduced by other routing methods under similar improvements, but with additional, lightweight operations introduced by the \emph{Avengers}.

\begin{table}[!t]
\small
\setlength{\tabcolsep}{3pt} 
\renewcommand{\arraystretch}{1.16}
\centering
\caption{Comparison of the \emph{Avengers} with baselines. \textit{Oracle$^*$} represents the best achievable score by selecting the optimal model per query. $^\dagger$Peak score on the test set. Max Expert represents the best performance of the ten models on the current dataset.}
\resizebox{\columnwidth}{!}{%
\begin{tabular}{lcccccccccccccccc}
\toprule
\multirow{2.5}{*}{\textbf{Setting}} &
  \multicolumn{3}{c}{\textbf{Mathematics}} &
  \multicolumn{2}{c}{\textbf{Code}} &
  \multicolumn{3}{c}{\textbf{Logical}} &
  \multicolumn{5}{c}{\textbf{Knowledge}} &
  \multicolumn{2}{c}{\textbf{Affective}} &
  \multirow{2.5}{*}{\textbf{Avg.}} \\ 
\cmidrule(lr){2-4}   
\cmidrule(lr){5-6}   
\cmidrule(lr){7-9}   
\cmidrule(lr){10-14} 
\cmidrule(lr){15-16} 
 &
  \textbf{AIME} &
  \textbf{M500.} &
  \textbf{LMB.} &
  \textbf{MBPP} &
  \textbf{HE.} &
  \textbf{KOR.} &
  \textbf{K\&K.} &
  \textbf{BBH} &
  \textbf{ARCC} &
  \textbf{MP.} &
  \textbf{GPQA} &
  \textbf{FinQA} &
  \textbf{MedQA} &
  \textbf{Emory.} &
  \textbf{MELD} &
   \\ 
   \midrule
\rowcolor{mylightgray}\multicolumn{17}{c}{\textbf{\textit{Small Langugage Model (Enhanced)}}} 
\\
\textbf{DS-Qwen}    & \cellcolor{toponeRed}61.67 & \cellcolor{toponeRed}93.00 & \cellcolor{toponeRed}65.71 & 56.67 & 43.90 & \cellcolor{toponeRed}52.00 & \cellcolor{toponeRed}76.10 & \cellcolor{toponeRed}80.00 & 88.82 & \cellcolor{toptwoRed}59.14 & \cellcolor{toponeRed}59.15 & \cellcolor{toptwoRed}68.53 & 40.85 & 29.84 & 38.64 & 60.93 \\

\textbf{Fin-R1} & 13.33 & \cellcolor{toptwoRed}80.20 & 35.00 & 69.30 & 77.44 & 37.52 & 18.14 & 68.24 & \cellcolor{toptwoRed}91.89 & 51.55 & 20.76 & \cellcolor{toponeRed}70.71 & 67.09 & \cellcolor{toptwoRed}41.18 & 57.55 & 53.33 \\

\textbf{Qwen-it}         & \cellcolor{toptwoRed}15.00 & 78.60 & \cellcolor{toptwoRed}37.14 & \cellcolor{toptwoRed}70.73 & \cellcolor{toponeRed}81.10 & \cellcolor{toptwoRed}49.60 & \cellcolor{toptwoRed}35.92 & 63.24 & 88.91 & 58.34 & 30.13 & 68.00 & 65.28 & 39.89 & \cellcolor{toptwoRed}57.87 & 55.98 \\

\textbf{Qwen-Coder}   & 11.67 & 75.60 & 35.00 & \cellcolor{toponeRed}76.39 & \cellcolor{toptwoRed}78.66 & 34.64 & 27.57 & 61.48 & 86.52 & 49.25 & 32.14 & 65.13 & 55.38 & 40.46 & \cellcolor{toponeRed}59.58 & 52.63 \\

\textbf{gemma-2-it}      & 1.67  & 54.00 & 22.14 & 62.32 & 64.02 & 34.32 & 15.71 & 63.52 & 89.42 & 53.55 & 34.38 & 66.26 & 66.46 & 39.60 & 54.38 & 48.12 \\

\textbf{glm-4-chat}      & 5.00  & 58.40 & 22.14 & 62.01 & 65.85 & 37.60 & 21.57 & 47.59 & \cellcolor{toponeRed}92.15 & 51.75 & 31.25 & 58.59 & 64.96 & \cellcolor{toponeRed}41.32 & 57.46 & 47.84 \\

\textbf{Llama-3.1-it}       & 1.67  & 49.80 & 20.71 & 61.81 & 71.95 & 27.60 & 11.71 & 65.74 & 88.48 & 47.35 & 25.67 & 53.97 & \cellcolor{toptwoRed}69.76 & 35.15 & 51.62 & 45.53 \\

\textbf{Granite-3.1-it}     & 1.67  & 61.40 & 20.00 & 37.17 & 39.63 & 31.44 & 19.29 & 36.39 & 85.24 & 44.06 & \cellcolor{toptwoRed}34.82 & 65.74 & 59.70 & 39.60 & 48.30 & 41.63 \\

\textbf{UltraMedical} & 1.67  & 76.20 & 15.71 & 52.77 & 58.54 & 18.72 & 20.14 & 40.46 & 85.67 & 43.76 & 19.87 & 60.68 & \cellcolor{toponeRed}72.90 & 31.28 & 43.34 & 42.78 \\

\textbf{cogito-v1}  & 3.33  & 59.60 & 22.86 & 51.33 & 70.12 & 44.48 & 28.00 & \cellcolor{toptwoRed}75.83 & 90.01 & \cellcolor{toponeRed}61.94 & 31.92 & 64.69 & 69.36 & 39.02 & 55.68 & 51.21 \\

\midrule
\textbf{Average} & 11.67 & 68.68 & 29.64 & 60.05 & 65.12 & 36.79 & 27.42 & 60.25 & 88.71 & 52.07 & 32.01 & 64.23 & 63.17 & 37.73 & 52.44 & 50.00 \\
\textbf{Max Expert} & 61.67 & 93.00 & 65.71 & 76.39 & 81.10 & 52.00 & 76.10 & 80.00 & 92.15 & 61.94 & 59.15 & 70.88 & 72.90 & 41.32 & 59.58 & 69.59 \\
\rowcolor{mylightgray}\textbf{\textit{Oracle$^*$}} & 
\textit{63.33} & 
\textit{96.60} & 
\textit{69.29} & 
\textit{89.43} & 
\textit{95.73} & 
\textit{65.76} & 
\textit{82.43} & 
\textit{96.85} & 
\textit{96.83} & 
\textit{86.41} & 
\textit{85.49} & 
\textit{85.44} & \textit{91.99} & \textit{66.86} & \textit{82.71} & \textit{83.68} \\

\midrule
\rowcolor{mylightgray}\multicolumn{17}{c}{\textbf{\textit{Proprietary Model Baseline}}} 
\\
\textbf{GPT-4o} & 10.00 & 76.00 &	35.71 &	82.64 &	85.36 &	57.68 &	32.57 &	79.53 &	93.86 & 59.84 & 44.42 &72.28 & 82.17 & 38.31 & 52.92 & 60.22 \\
\textbf{GPT-4.5-Preview} & 31.67 & 88.60 & 	50.00 &	86.69&	69.51&	57.44&	89.29&	91.36&	94.60&	59.94&	51.78&	71.75&	92.85&	39.45&	47.84& 	68.18\\
\textbf{GPT-4.1}  & 45.00 & 89.20 & 52.14 & 57.70 & 92.07 & 48.48 & 84.43 & 90.74 & 95.39 & 65.13 & 62.05 & 71.05 & 88.77 & 39.45 & 56.33 & 69.20 \\ 
\midrule
\rowcolor{mylightgray}\multicolumn{17}{c}{\textbf{\textit{Mixture-based Baseline}}} \\
\textbf{MoA} & 23.33 & 82.67 & 33.33 & 54.61 & 84.51 & 49.87 & 34.28 & 47.53 & 91.81 & 64.54 & 50.37 & 71.04 & 71.20 & 39.17 & 52.76 & 56.70 \\
\textbf{MoA (Oracle)} & 27.77 & 84.67 & 54.76 & 69.40 & 83.54 & 51.28 & 31.14 & 57.50 & 92.75 & 61.74 & 50.44 & 70.71 & 76.05 & 39.45 & 51.95 & 60.21 \\
\textbf{SMoE} & 43.00 & 88.20 & 42.86 & 65.81 & 70.12 & 48.88 & 43.71 & 67.69 & 87.63 & 58.94 & 43.30 & 65.21 & 64.96 & 37.30 & 51.38 & 58.60 \\
\midrule

\rowcolor{mylightgray}\multicolumn{17}{c}{\textbf{\textit{Router-based Baseline (Enhanced)}}} \\
\textbf{Random Router}  & 10.00 & 66.20 & 32.86 & 61.29 & 62.80 & 35.60 & 27.57 & 61.20 & 88.65 & 51.25 & 33.04 & 64.60 & 62.61 & 38.16 & 50.89 & 49.78 \\
\textbf{LLM Router} & 61.67 & 91.80 & 64.29 & 69.92 & 78.05 & 37.12 & 74.71 & 80.28 & 88.99 & 60.14 & 45.31 & 66.78 & 73.84 & 32.71 & 45.78 & 64.76 \\
\textbf{RouterDC$^\dagger$} & 58.89 & 89.87 & 64.76 & 68.53 & 73.20 & 44.64 & 73.33 & 73.33 & 90.00 & 54.68 & 43.41 & 68.12 & 66.81 & 42.00 & 58.87 & 64.70 \\
\textbf{MODEL-SAT$^\dagger$} & 61.11 & 91.73 & 65.24 & 69.42 & 80.00 & 54.77 & 72.76 & 82.96 & 90.34 & 65.12 & 58.67 & 68.70 & 66.70 & 40.19 & 58.00 & 68.38 \\
\textbf{EmbedLLM$^\dagger$} & 61.11 & 90.80 & 64.76 & 74.27 & 78.80 & 55.73 & 73.81 & 85.06 & 91.14 & 65.32 & 57.33 &	67.12 & 71.83 & 41.33 & 60.22 & 69.24 \\
\midrule
\bfseries The \emph{Avengers} (ours) & 
\bfseries 61.67 & \bfseries 92.89 & \bfseries 65.71 & \bfseries 76.08 & \bfseries 84.86 &
\bfseries 55.95 & \bfseries 76.14 & \bfseries 84.07 & \bfseries 92.39 & \bfseries 66.21 &
\bfseries 55.76 & \bfseries 71.53 & \bfseries 73.89 & \bfseries 41.38 & \bfseries 59.61 & 
\bfseries 70.54 \\


\scriptsize\itshape - vs GPT-4.1 (\%) &
\scriptsize\textcolor{red}{↑37.04} & 
\scriptsize\textcolor{red}{↑4.14} & 
\scriptsize\textcolor{red}{↑26.03} &
\scriptsize\textcolor{red}{↑31.86} & 
\scriptsize\textcolor{myForestGreen}{↓7.83} & 
\scriptsize\textcolor{red}{↑15.41} & 
\scriptsize\textcolor{myForestGreen}{↓9.82} & 
\scriptsize\textcolor{myForestGreen}{↓7.35} & 
\scriptsize\textcolor{myForestGreen}{↓3.15} & 
\scriptsize\textcolor{red}{↑1.66} & 
\scriptsize\textcolor{myForestGreen}{↓10.14} 
& \scriptsize\textcolor{red}{↑0.67} & 
\scriptsize\textcolor{myForestGreen}{↓16.76} &
\scriptsize\textcolor{red}{↑4.90} & 
\scriptsize\textcolor{red}{↑5.83} & 
\scriptsize\textcolor{red}{↑1.95} 
\\

\scriptsize\itshape - vs MoA (Oracle) (\%) &
\scriptsize\textcolor{red}{↑122.06} & \scriptsize\textcolor{red}{↑9.71} & 
\scriptsize\textcolor{red}{↑20.00} &
\scriptsize\textcolor{red}{↑9.63} & \scriptsize\textcolor{red}{↑1.58} & 
\scriptsize\textcolor{red}{↑9.11} &  \scriptsize\textcolor{red}{↑144.52} & 
\scriptsize\textcolor{red}{↑46.21} & \scriptsize\textcolor{myForestGreen}{↓0.39} & 
\scriptsize\textcolor{red}{↑7.24} & \scriptsize\textcolor{red}{↑10.54} & 
\scriptsize\textcolor{red}{↑1.16} &
\scriptsize\textcolor{myForestGreen}{↓2.84} & \scriptsize\textcolor{red}{↑4.90} & \scriptsize\textcolor{red}{↑14.75} & \scriptsize\textcolor{red}{↑17.16}
\\

\scriptsize\itshape - vs EmbedLLM (\%)  &
\scriptsize\textcolor{red}{↑0.91} & 
\scriptsize\textcolor{red}{↑2.31} &
\scriptsize\textcolor{red}{↑1.47} &
\scriptsize\textcolor{red}{↑2.44} & 
\scriptsize\textcolor{red}{↑7.69} &
\scriptsize\textcolor{red}{↑0.39} &
\scriptsize\textcolor{red}{↑3.16} &
\scriptsize\textcolor{myForestGreen}{↓1.16} &
\scriptsize\textcolor{red}{↑1.37} &
\scriptsize\textcolor{red}{↑1.37} &
\scriptsize\textcolor{myForestGreen}{↓2.75} &
\scriptsize\textcolor{red}{↑6.57} &
\scriptsize\textcolor{red}{↑2.87} &
\scriptsize\textcolor{red}{↑0.12} &
\scriptsize\textcolor{myForestGreen}{↓1.00} &
\scriptsize\textcolor{red}{↑1.88} 
\\

\bottomrule
\end{tabular}%
}
\label{tab:main-result}
\end{table}

\begin{table}[!t]
\small
\centering
\caption{Out-of-Distribution performance. $^\dagger$Peak score on the test set.}
\renewcommand{\arraystretch}{0.9}
\resizebox{0.9\columnwidth}{!}{%
\begin{tabular}{lcccccc}
\toprule
\textbf{Setting}         
& \textbf{MathBench} 
& \textbf{StudentEval}
& \textbf{Winogrande}
& \textbf{BrainTeaser}
& \textbf{DailyDialog} 
& \textbf{Avg.}  
\\ \midrule
\rowcolor{mylightgray}\multicolumn{7}{c}{\textit{\textbf{Small Langugage Model (Enhanced)}}} \\
\textbf{DS-Qwen}         & \cellcolor{toponeRed}97.33     & 53.83       & 65.98      & 68.46       & 38.30       & 64.78 \\
\textbf{Fin-R1}          & 65.33     & \cellcolor{toptwoRed}65.69       & \cellcolor{toponeRed}77.82      & 70.13       & 43.30       & 64.45 \\
\textbf{Qwen-it}         & \cellcolor{toptwoRed}83.33     & 65.48       & 69.85      & 67.78       & 53.30       & 67.95 \\
\textbf{Qwen-Coder}      & 75.33     & 63.80       & 67.56      & 59.84       & \cellcolor{toptwoRed}57.70       & 64.85 \\
\textbf{gemma-2-it}      & 53.30     & 64.22       & 73.16      & \cellcolor{toponeRed}76.00       & 50.90       & 63.52 \\
\textbf{glm-4-chat}     & 56.67     & 60.65       & \cellcolor{toptwoRed}77.74      & 62.68       & \cellcolor{toponeRed}59.10       & 63.37 \\
\textbf{Llama-3.1-it}    & 49.33     & \cellcolor{toponeRed}68.84       & 64.09      & 70.03       & 37.70       & 58.00 \\
\textbf{Granite-3.1-it}  & 57.33     & 40.50       & 72.77      & 61.80       & 35.20       & 53.52 \\
\textbf{UltraMedical}    & 38.67     & 57.61       & 62.43      & 62.88       & 32.60       & 50.84 \\
\textbf{cogito-v1}       & 66.00     & 56.45       & 64.09      & \cellcolor{toptwoRed}73.16       & 37.70       & 59.48 \\
\midrule
\textbf{Average}         & 64.26     & 59.71       & 69.55      & 67.28       & 44.58       & 61.07 \\
\textbf{Max Expert}      & 97.33     & 68.84       & 77.82      & 76.00       & 59.10       & 67.95 \\
\rowcolor{mylightgray}
\textbf{\textit{Oracle}} & 
\textit{100.00}  &  \textit{84.89}  &  \textit{95.26}   &
\textit{94.32}   &  \textit{75.30}  &  \textit{89.95} \\
\midrule
\rowcolor{mylightgray}\multicolumn{7}{c}{\textit{\textbf{Router Baseline (Enhanced)}}} \\
\textbf{Random Router}  & 64.67   & 55.82  & 69.53   & 66.70 & 43.60  & 60.06 \\
\textbf{LLM Router} & 94.67 & 67.79 & 65.27 & 67.48 & 37.10 & 66.46 \\
\textbf{RouterDC$^\dagger$} & 77.27 & 62.50 & 69.77 & 70.91 & 59.20 & 66.89 \\
\textbf{MODEL-SAT$^\dagger$} & 97.20 & 58.22 & 68.37 & 70.58 & 48.40 & 68.55 \\
\textbf{EmbedLLM$^\dagger$} & 97.20 & 63.06 & 71.11 & 69.34 & 47.98 & 69.77 \\
\midrule
\bfseries The \emph{Avengers} (ours) & 96.67     & 68.52       & 77.43      & 71.79       & 57.70       & 74.42 \\ 

\scriptsize\itshape - vs EmbedLLM (\%) &
\scriptsize\textcolor{myForestGreen}{↓0.55} &
\scriptsize\textcolor{red}{↑4.98} & 
\scriptsize\textcolor{red}{↑19.17} &
\scriptsize\textcolor{red}{↑2.20} &
\scriptsize\textcolor{red}{↑24.30} &
\scriptsize\textcolor{red}{↑8.14} \\
\bottomrule
\end{tabular}%
}
\label{tab:ood-result}
\end{table}


\subsection{The Key Elements that Make the \emph{Avengers} Work}
To understand why the \emph{Avengers} is both simple and effective, we conduct targeted ablations on four design choices—\textbf{embedding model}, \textbf{clustering method}, \textbf{model selection}, and \textbf{ensemble strategy}—while keeping all other components fixed.

\begin{table*}[!htbp]
    \centering
    \begin{minipage}[t]{0.30\linewidth}
    \centering
    \caption{Average score of the \emph{Avengers} when equipped with different embedding models.}
    \small
    \resizebox{\linewidth}{!}{%
        \begin{tabular}{@{}lcc@{}}
        \toprule
        \textbf{Embed.} & \textbf{Para./Dim.} & \textbf{Avg.} \\
        \midrule
        bge-m3  & 0.56/1\,024 & 70.28 \\
        text-3-small$^{\dagger}$   & -\-/1\,536 & 70.42 \\
        gte-q2-1.5B & 1.5/1\,536 & 70.48 \\
        text-3-large$^{\dagger}$ & -/3\,072 & \textbf{70.70} \\
        gte-q2-7B & 7/3\,584 & \underline{70.54} \\
        \bottomrule
        \end{tabular}
    }
    \label{tab:diff-embedding-model}
    \end{minipage}
    \hfill
    \begin{minipage}[t]{0.26\linewidth}
    \centering
    \caption{Comparison of average scores for different clustering methods.}
    \small  
    \resizebox{\columnwidth}{!}{%
    \begin{tabular}{@{}lc@{}}
        \toprule
        \bfseries Cluster Method & \bfseries Avg. \\ 
        \midrule
        K-Means &   70.54  \\
        Hierarchial Clustering &  \textbf{70.71} \\
        GMM &  70.58       \\
        Spectral Clustering & \underline{70.61} \\
        BIRCH     &  70.04    \\ 
        \bottomrule
        \end{tabular}%
    }
    \label{tab:cluster}
    \end{minipage}
    \hfill
    \begin{minipage}[t]{0.28\linewidth}
    \centering
    \small
    \caption{Ablation on the \emph{Avengers} ensemble strategies (excluding code tasks).}
    \resizebox{\columnwidth}{!}{%
        \begin{tabular}{@{}lc@{}}
        \toprule
        \textbf{Ensemble Strategy} & \textbf{Avg.} \\
        \midrule
        Direct (CoT)   & 59.19 \\
        Aggregation &  56.73 \\
        Model-Switch  & \underline{66.42} \\
        Self-Consistency   & \textbf{66.98} \\
        \bottomrule
        \end{tabular}
    }
    \label{1234}
    \end{minipage}
\end{table*}

\paragraph{Embedding Model}

Since clustering quality directly depends on the embedding model’s semantic representation, we investigate how the choice of embedding model affects the performance of the \emph{Avengers}. Specifically, we compare five embedding models spanning different parameter sizes and embedding dimensions (Table~\ref{tab:diff-embedding-model}), including \textbf{bge-m3}~\citep{bge-m3}, \textbf{text-embedding-3-small} (text-3-small)~\citep{openai_te3_2024}, \textbf{gte-qwen2-1.5B-instruct} (gte-q2-1.5B)~\citep{li2023gteqwen}, \textbf{text-embedding-3-large} (text-3-large)~\citep{openai_te3_2024} and \textbf{gte-qwen2-7B-instruct} (gte-q2-7B)~\citep{li2023gteqwen}. Results show minimal performance variation (70.28 $\rightarrow $ 70.70), demonstrating that the proposed method is robust to changes in embedding model selection.


\paragraph{Clustering Method}
Since the \emph{Avengers} relies on query clustering, the choice of clustering method could potentially impact overall performance. To assess this influence, we conduct experiments using five classical clustering algorithms: \textbf{K-Means}~\citep{lloyd1982least, macqueen1967some}, \textbf{Hierarchical Clustering}~\citep{ward1963hierarchical}, \textbf{Gaussian Mixture Models} (GMM)~\citep{dempster1977maximum}, \textbf{Spectral Clustering}~\citep{von2007tutorial} and \textbf{BIRCH}~\citep{zhang1996birch}. For all experiments, we fix the number of clusters at $K=64$ (details in Appendix~\ref{sec:cluster_detailed}). 
As shown in Table~\ref{tab:cluster}, performance varies minimally (70.04–70.71), indicating that the \emph{Avengers} framework is robust and stable across different clustering algorithms.

\begin{figure}[!htbp]     
\centering 
    \begin{minipage}[t]{0.48\textwidth} 
        \centering
        \includegraphics[width=\linewidth]{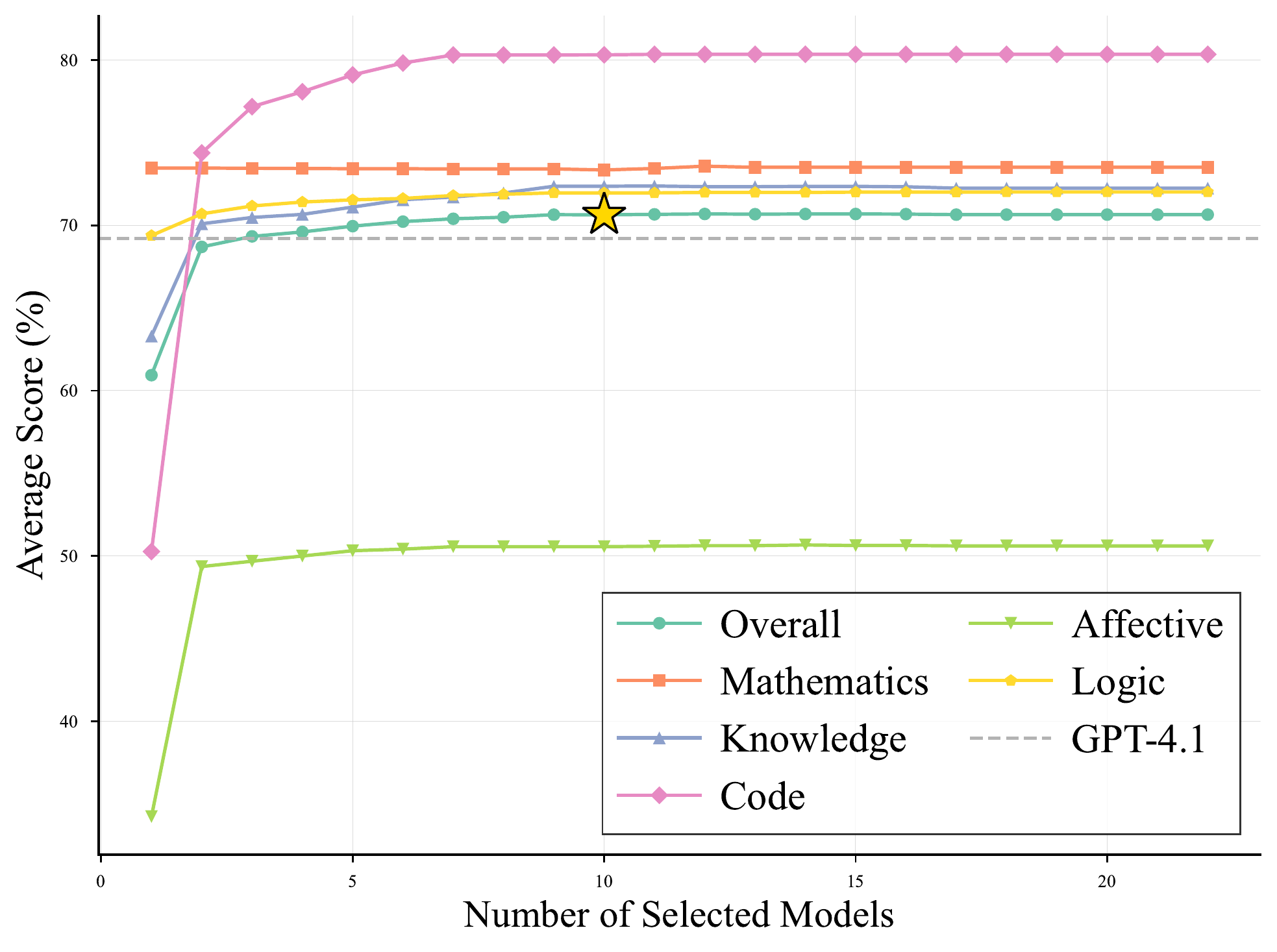} 
        \caption{Impact of selected models' number.}
        \label{fig:modelselection_mp} 
    \end{minipage}
    \hfill 
    \begin{minipage}[t]{0.48\textwidth}
        \centering
        \includegraphics[width=\linewidth]{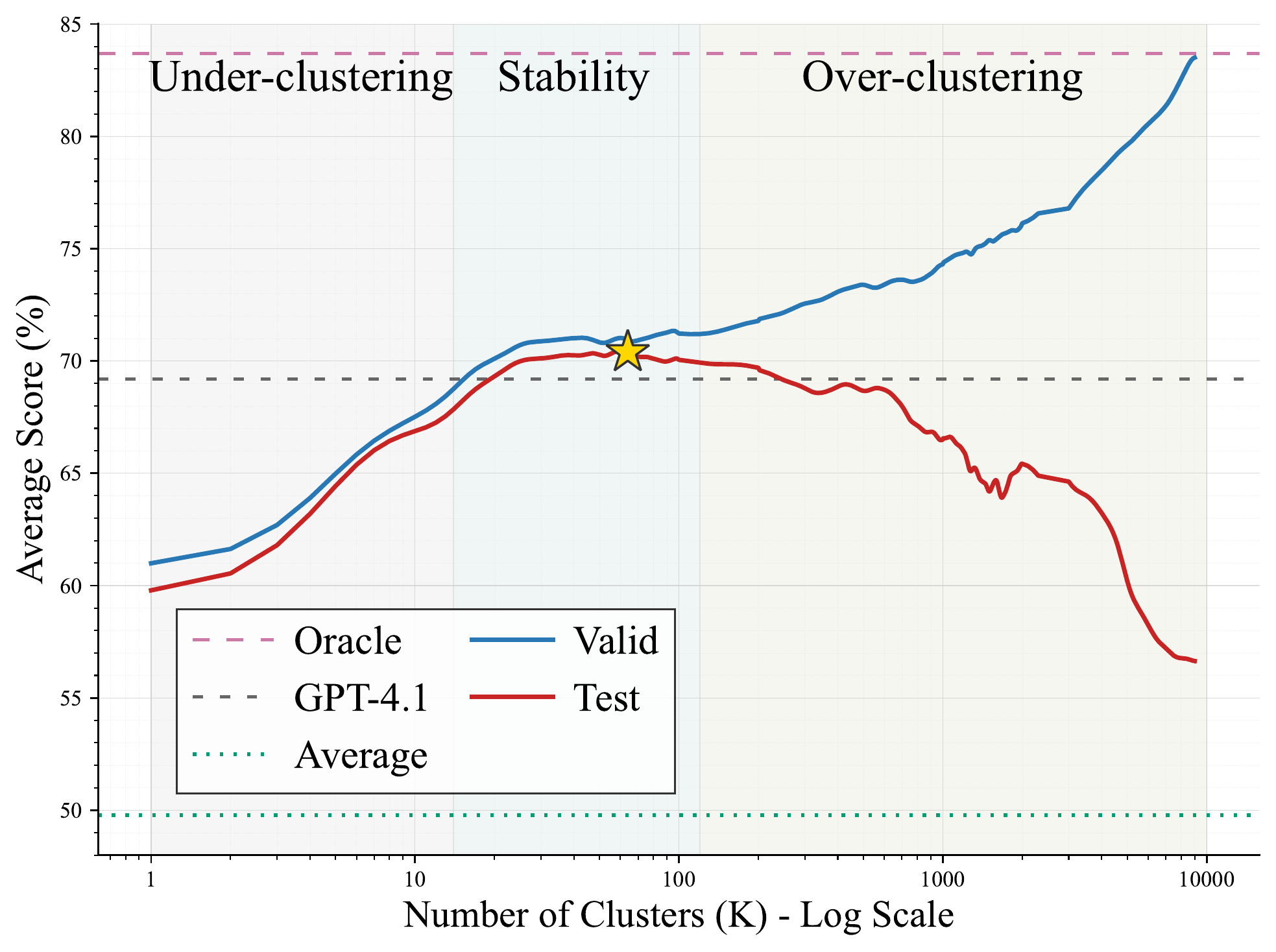}
        \caption{Impact of cluster count $K$.} %
        \label{fig:k_impact}
    \end{minipage}
      \vspace{-2pt}
\end{figure}

\paragraph{Model Selection}
\label{sec:model-selection}
Figure \ref{fig:modelselection_mp} shows the performance as the number of selected models in the \emph{Avengers} increases from 1 to 22. The overall performance rises sharply, matching GPT-4.1 with only 3 models, and plateaus at approximately 10 models. Knowledge, code, and affective tasks exhibit similar trends, while mathematics and logic tasks gain little due to a single model’s dominance. These results confirm that the \emph{Avengers} effectively leverages model complementarity, achieving strong general-purpose performance without needing to deploy all available models.

\paragraph{Ensemble Strategy}

To analyze ensemble strategies within the \emph{Avengers}, we compare Direct (CoT), Aggregation, Model-Switch, and Self-Consistency. Among these, Self-Consistency achieves the best overall performance due to its simplicity, effectiveness, and robustness across tasks.

\paragraph{The Impact of Cluster Count $K$}
Figure~\ref{fig:k_impact} demonstrates the performance dynamics of the \emph{Avengers} on both validation and test sets as the cluster count $K$ varies from 1 to 10,000 (logarithmic scale on the horizontal axis). Notably, our method surpasses GPT-4.1 across a wide range of $K$ values (approximately 14 to 140), indicating robustness and minimal sensitivity to the choice of $K$.

\section{Conclusion}
In this paper, we introduce the \emph{Avengers}, a simple yet effective framework to unite multiple smaller language models (SLMs) and challenge the dominance of proprietary large models. The core of the \emph{Avengers} involves straightforward embedding, clustering, scoring, and voting, without requiring neural network training, prompt engineering, or careful architecture-specific model choices. Moreover, it is highly adaptable to new domains and new models, and supports automatic model set construction. 

Experimental results show that by combining 10 SLMs (each with approximately 7B parameters), the \emph{Avengers} achieves highly competitive performance across diverse tasks. Specifically, it surpasses OpenAI’s flagship model, GPT-4.1, on nine out of 15 diverse datasets spanning mathematics, coding, logical reasoning, knowledge, and affective recognition tasks. Notably, it achieves an 18.21\% improvement over GPT-4.1 on mathematics tasks and a 7.46\% improvement on coding tasks. Additionally, the \emph{Avengers} demonstrates superior out-of-distribution (OOD) generalization, outperforming state-of-the-art router-based methods by at least 8\%.
Further analyses reveal the robustness of the \emph{Avengers} across different clustering methods, embedding model choices, and key hyperparameters such as the number of clusters ($k$). With an increasing number of integrated models, the \emph{Avengers} exhibits complementary advantages, especially in knowledge, code and affective tasks. 

The superior performance of the \emph{Avengers} highlights the untapped potential of community-driven, open-source models and provides a promising pathway toward democratizing the advancements of LM research.

\bibliography{references} 
\bibliographystyle{unsrtnat}



\appendix
\section{Technical Appendices and Supplementary Material}

\subsection{Further Related Work}
\label{sec: app_related_work}

\paragraph{Mixture-based methods} This line of research typically processes each query using multiple models in parallel to generate several candidate responses, and then synthesizes these into a final output using an aggregator model. 
Mixture-of-Agents (MoA)~\citep{wang2024mixture}, a notable example, surpasses GPT-4 Omni by leveraging iterative interactions among multi-level agents composed of a pool of medium-sized models (over 70B parameters). 
Sparse-MOA~\citep{li2024smoa} leverages response selection and early stopping to sparsify information flow, utilizes role-playing for enhanced diversity, achieving comparable performance to standard MoA at a reduced cost. Self-MoA~\citep{li2025rethinking} demonstrates that prioritizing high-quality outputs from the model is crucial, and shows that selecting the single best-performing model on the current task to serve simultaneously as both Aggregator and Proposer can surpass the standard MoA. Symbolic-MoE~\citep{chen2025symbolicmixtureofexpertsadaptiveskillbased} introduces a skill-based adaptive routing strategy that leverages an additional SLM to generate skill profiles for each query and candidate SLM, dynamically recruiting the most suitable expert models at the query level. 

 Compared to router-based approaches, mixture-based methods are training-free and often demonstrate greater performance gains over single models. However, such frameworks heavily depend on hand-crafted prompts and meticulously selected architectural components, such as the careful selection of aggregator models. Moreover, they often entail intricate hybrid (serial-parallel) execution flows, leading to substantial inference overhead. Last but not least, they necessitate the abilities of instruction following and prompt robustness---which are key shortcomings of small language models. Although the \emph{Avengers} does \textit{not} fall within this paradigm, we include comparisons with representative mixture-based methods to highlight our framework's effectiveness, especially in the context of small language models.



\paragraph{Merging-based methods}
This line of research combines the parameters of several fine-tuned models (typically derived from the same base model) to produce a single, more versatile model with enhanced task-specific performance and broader generalization.
Relevant works include TIES-Merging~\citep{yadav2023ties}, DARE~\citep{yu2024language}, Model Swarms~\citep{feng2024model}, LoRAHub~\citep{huanglorahub}, and GENOME~\citep{zhang2025nature}. Notably, recent studies predominantly focus on lightweight parameter merging techniques such as LoRA~\citep{hulora}, with limited exploration of full-parameter merging strategies, thereby constraining the performance ceiling. Furthermore, most existing methods assume architectural homogeneity~\citep{yadav2024matters}, i.e., homogeneous models derived from the same base model. This assumption is fundamentally different from ours. Thus, in this work, we do not compare our framework to this line of research.


\subsection{Models}
\label{sec:model}

Table~\ref{tab:model} shows the 10 mainly models used in the main experiments in our paper.
\begin{table}[!htbp]
\small
\centering
\caption{Detailed information of the models, including their abbreviations.} 
\begin{tabular}{@{}lcc@{}}
\toprule
\textbf{Model} & \textbf{Abbreviation} & \textbf{Parameters (B)} \\ 
\midrule
Fin-R1~\citep{liu2025finr1largelanguagemodel} & Fin-R1 & 7.61 \\
Qwen2.5-7B-Instruct~\citep{yang2024qwen2} & Qwen-it & 7.61 \\
Qwen2.5-Coder-7B-Instruct~\citep{hui2024qwen25codertechnicalreport} & Qwen-Coder & 7.61 \\
DeepSeek-R1-Distill-Qwen-7B~\citep{guo2025deepseek} & DS-Qwen & 7.61 \\
cogito-v1-preview-llama-8B & cogito-v1 & 8.03 \\
Llama-3.1-8B-Instruct~\citep{grattafiori2024llama} & Llama-3.1-it & 8.03 \\
Llama-3.1-8B-UltraMedical~\citep{zhang2024ultramedical} & UltraMedical & 8.03 \\
Granite-3.1-8B-Instruct & Granite-3.1-it & 8.03 \\
gemma-2-9b-it~\citep{team2024gemma} & gemma-2-it & 9.24 \\
glm-4-9b-chat~\citep{glm2024chatglm} & glm-4-chat & 9.40 \\
\midrule
\textbf{Average Parameters} & -- & 8.12 \\
\bottomrule
\end{tabular}
\label{tab:model} 
\end{table}

Table~\ref{tab:model2} presents the remaining 12 models utilized in the experiments detailed in Section~\ref{sec:model-selection}.
Notably, we begin our experiments with an initial pool consisting of all 22 models. Subsequently, an automated model selection method is applied to select the 10 models listed in Table~\ref{tab:model}. All results report in the main text, including those for baseline comparisons, consistently use the same set of models selected by the \emph{Avengers}.

\begin{table}[!htbp]
\small
\centering
\caption{Other models.} 
\begin{tabular}{@{}lc@{}}
\toprule
\textbf{Model}  & \textbf{Parameters (B)} \\ 
\midrule
Phi-4-mini-instruct~\citep{microsoft2025phi4minitechnicalreportcompact}  & 3.8 \\
Falcon3-7B-Instruct~\citep{Falcon3}  & 7 \\
mistral-7b-instruct-v0.3~\citep{jiang2023mistral7b}  & 7.25 \\
OLMo-2-1124-7B-Instruct~\citep{olmo20242olmo2furious}  & 7.3 \\
Eurus-2-7B-PRIME~\citep{cui2025process} & 7.61 \\
Qwen2.5-7B-MATH-Instruct~\citep{yang2024qwen2} & 7.61 \\
internlm3-8b-instruct~\citep{cai2024internlm2}  & 8 \\
Hermes-3-Llama-3.1-8B~\citep{teknium2024hermes3technicalreport}  & 8.03 \\
Medreason-8B~\citep{wu2025medreasonelicitingfactualmedical} & 8.03 \\
Llama-3.1-Nemotron-Nano-8B-v1~\citep{bercovich2025llamanemotronefficientreasoningmodels}  & 8.03 \\
Llama-3.1-Tulu-3.1-8B~\citep{lambert2024tulu3} & 8.03 \\
Yi-1.5-9B-Chat~\citep{ai2025yiopenfoundationmodels}  & 8.83 \\
\midrule
\textbf{Average Parameters} & 7.46 \\
\bottomrule
\end{tabular}
\label{tab:model2} 
\end{table}

\subsection{Datasets}
\label{app:dataset}

We provide the detailed information of datasets in Table~\ref{tab:benchmarks}.

\begin{table}[!htbp]
\small
\centering
\caption{Detailed information of the datasets.}
\begin{tabular}{@{}llll@{}}
\toprule
\textbf{Dataset} & \textbf{Category} & \textbf{Metrics} & \textbf{Size} \\ \midrule
AIME          & Mathematics         & Accuracy, 0-shot & 60  \\
MATH500~\citep{lightman2023lets}       & Mathematics         & Accuracy, 0-shot & 500 \\
LiveMathBench~\citep{liu2024livemathbench} & Mathematics         & Accuracy, 0-shot & 140 \\
MBPP~\citep{austin2021mbpp}          & Code     & Pass@1, 0-shot   & 974 \\
HumanEval~\citep{chen2021evaluating}     & Code     & Pass@1, 0-shot   & 164 \\
KORBench~\citep{ma2024korbenchbenchmarkinglanguagemodels}      & Logic   & Accuracy, 3-shot  & 1\,250 \\
Knights and Knaves~\citep{xie2024memorization} & Logic & Accuracy, 0-shot & 700 \\
BBH~\citep{suzgun2022challenging}           & Logic   & Accuracy, 3-shot & 1\,080  \\
ARC Challenge~\citep{clark2018think} & Knowledge   & Accuracy, 0-shot & 1\,172 \\
MMLUPro~\citep{wang2024mmlupro}   & Knowledge   & Accuracy, 0-shot & 1\,000  \\
GPQA~\citep{rein2024gpqa}   & Knowledge   & Accuracy, 0-shot & 448  \\
FinQA~\citep{chen2021finqa}         & Knowledge  & Accuracy, 0-shot & 1\,147  \\
MedQA~\citep{jin2021disease}         & Knowledge  & Accuracy, 0-shot & 1\,273 \\
EmoryNLP~\citep{byrkjeland2018ternary}      & Affective & Accuracy, 0-shot & 697 \\ 
MELD~\citep{poria2019meldmultimodalmultipartydataset}      & Affective & Accuracy, 0-shot & 1\,232  \\ 
\midrule
\multicolumn{4}{c}{\cellcolor{mylightgray}\textit{\textbf{Out of Distribution Datasets}}} \\
MathBench~\citep{liu2024mathbench}    & Mathematics & Accuracy, 0-shot & 150 \\
StudentEval~\citep{babe2023studenteval} & Code & Pass@1, 0-shot & 953 \\
Winogrande~\citep{sakaguchi2021winogrande}  & Logic & Accuracy, 0-shot & 1\,267 \\
BRAINTEASER~\citep{rahimi2024nimz} & Knowledge & Accuracy, 0-shot & 1\,021 \\
DailyDialog~\citep{li2017dailydialog} & Affective & Accuracy, 0-shot & 1\,000 \\
\bottomrule
\end{tabular}%
\label{tab:benchmarks}
\end{table}








\subsection{Baseline}
\label{baseline}
\paragraph{LLM Router}
We use Qwen2.5-7B-Instruct as the router, which takes the current query along with descriptions (profiles) of available models to automatically select one or multiple suitable models for the query. This router requires no additional training; it only relies on natural-language model profiles prepared for each model.
In our implementation, we introduce an automatic method to generate these model profiles. Specifically, we first collect the performance scores of all models on the validation set. Then, using task descriptions and these scores, we prompt Qwen2.5-7B-Instruct to automatically generate the corresponding model profiles. To ensure consistent performance, we keep the request parameters fixed at $\text{temperature}=0.2$ and $\text{top}\_p=1.0$ for both profile generation and inference in the LLM router.

\begin{figure}[htbp]
    \centering
    \includegraphics[width=0.85\linewidth]{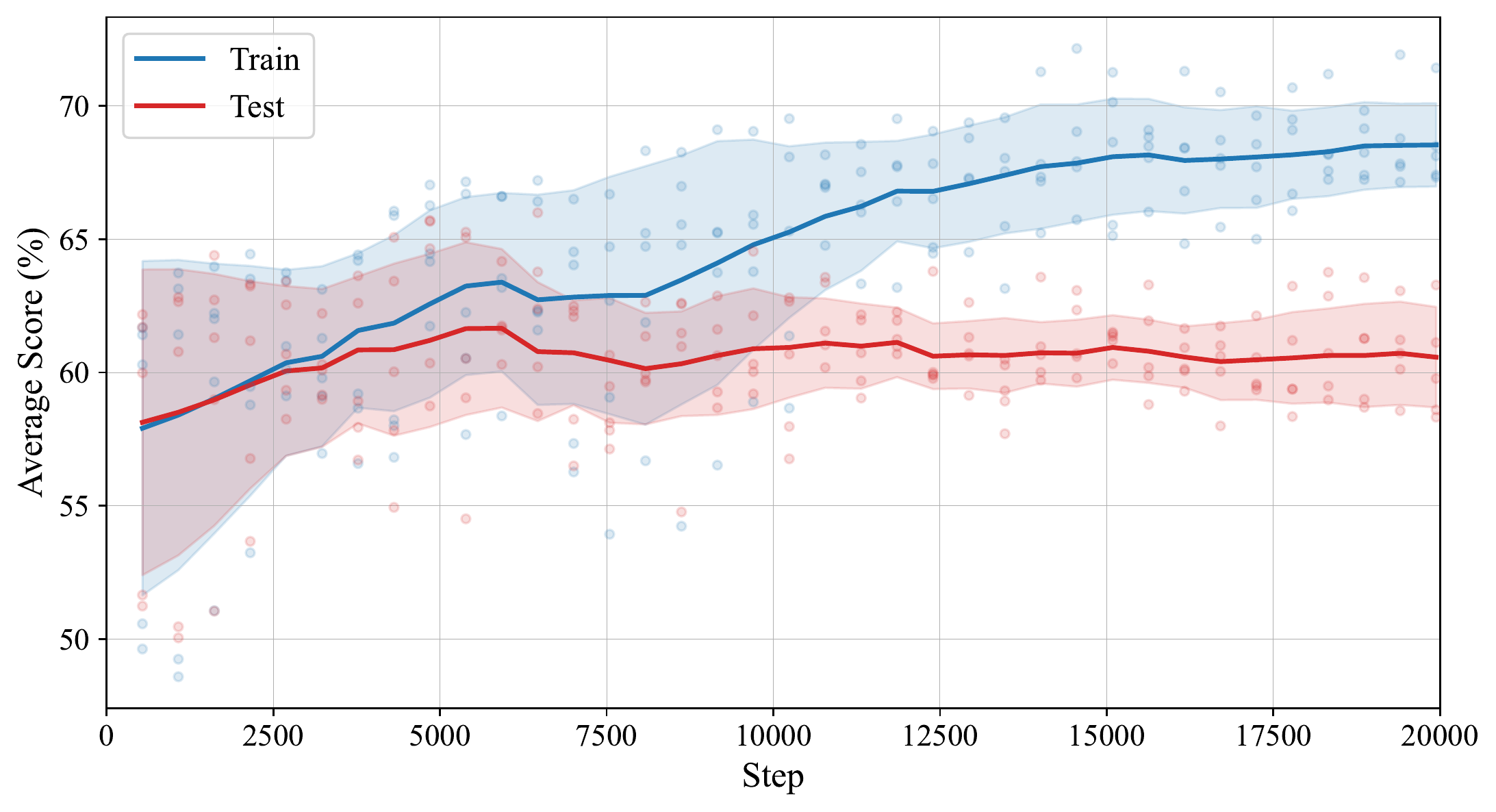}
    \caption{The training and test accuracy curves on RouterDC.}
    \label{fig:routerdc_acc}
\end{figure}

\paragraph{RouterDC}
We employ the official implementation~\citep{chen2024routerdc}, substituting the encoder with gte-qwen2-7B-instruct to ensure a fair comparison with our method. We utilize DeepSpeed for distributed multi-GPU training on eight NVIDIA A800-80G GPUs, setting the batch size per GPU to 8 while keeping all other hyperparameters unchanged. Our implementation comprises approximately 7B trainable parameters. The accuracy curves for both training and test are shown in Figure \ref{fig:routerdc_acc}.

\begin{figure}[htbp]
    \centering
    \includegraphics[width=0.85\linewidth]{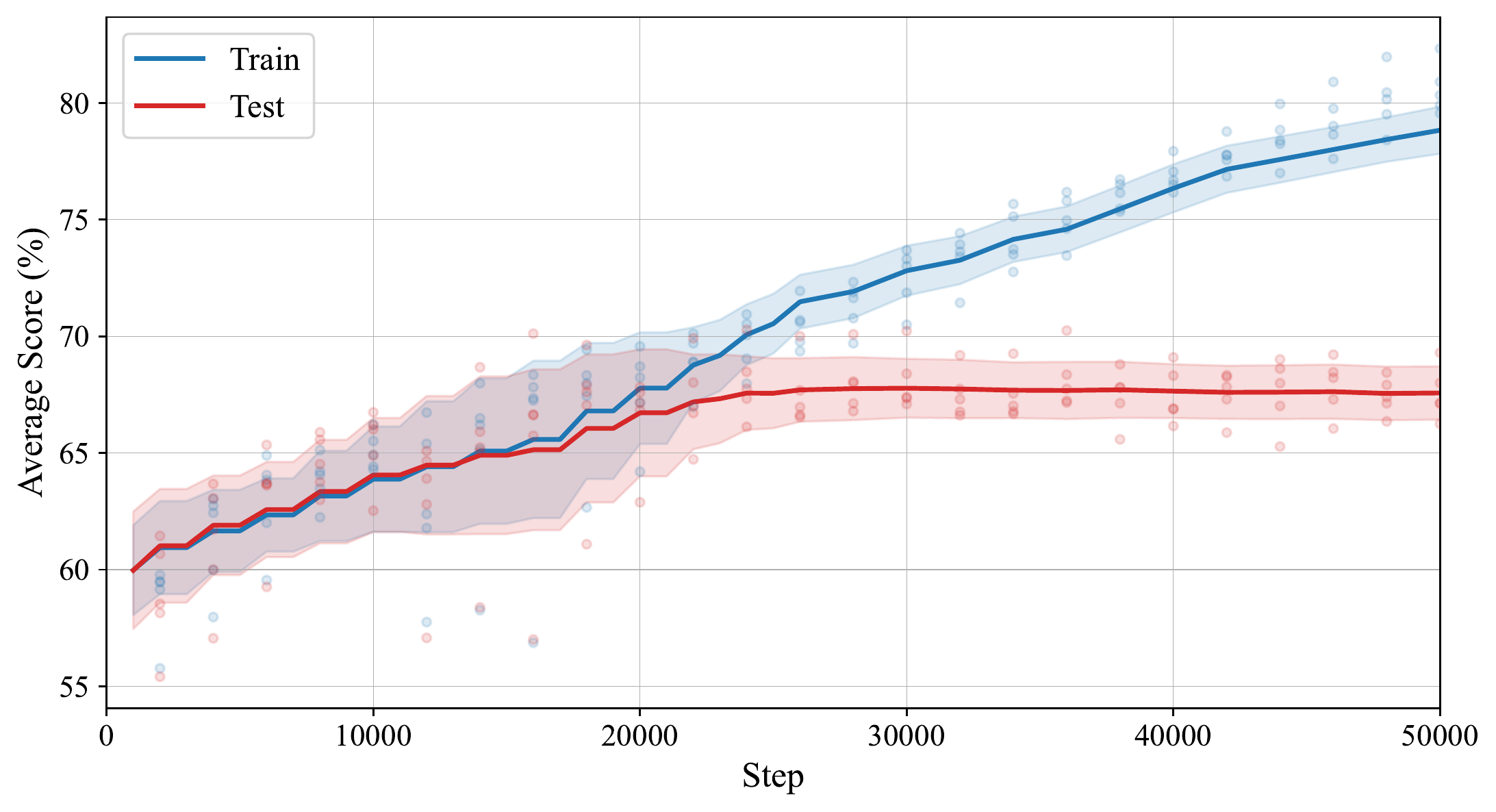}
    \caption{The training and test accuracy curves on MODEL-SAT.}
    \label{fig:model_sat_acc}
\end{figure}

\paragraph{MODEL-SAT} 
Since the official implementation provided by \citep{zhang2025capability} is incomplete, we re-implement the primary method, reproduce the code, and report the results. We employ gte-qwen2-7B-instruct as the embedding model and Qwen2.5-7B-Instruct as the language model, connecting them via a two-layer MLP projector. For efficient training, we leverage DeepSpeed for distributed multi-GPU training across eight NVIDIA A800-80GB GPUs, with a batch size of 2 per GPU. The learning rate is configured at 2e-6 for both the embedding and language models, while the projector uses a higher learning rate of 1e-4. Initially, we exclusively fine-tune the projector for approximately 10,000 steps. Following this phase, we continue fine-tuning all model parameters for the remainder of the training process. A warmup ratio of 0.1 is applied to stabilize the early training stages. Our implementation comprises approximately 14B trainable parameters. Figure \ref{fig:model_sat_acc} illustrates the training and test accuracy curves.

\begin{figure}[htbp]
    \centering
    \includegraphics[width=0.85\linewidth]{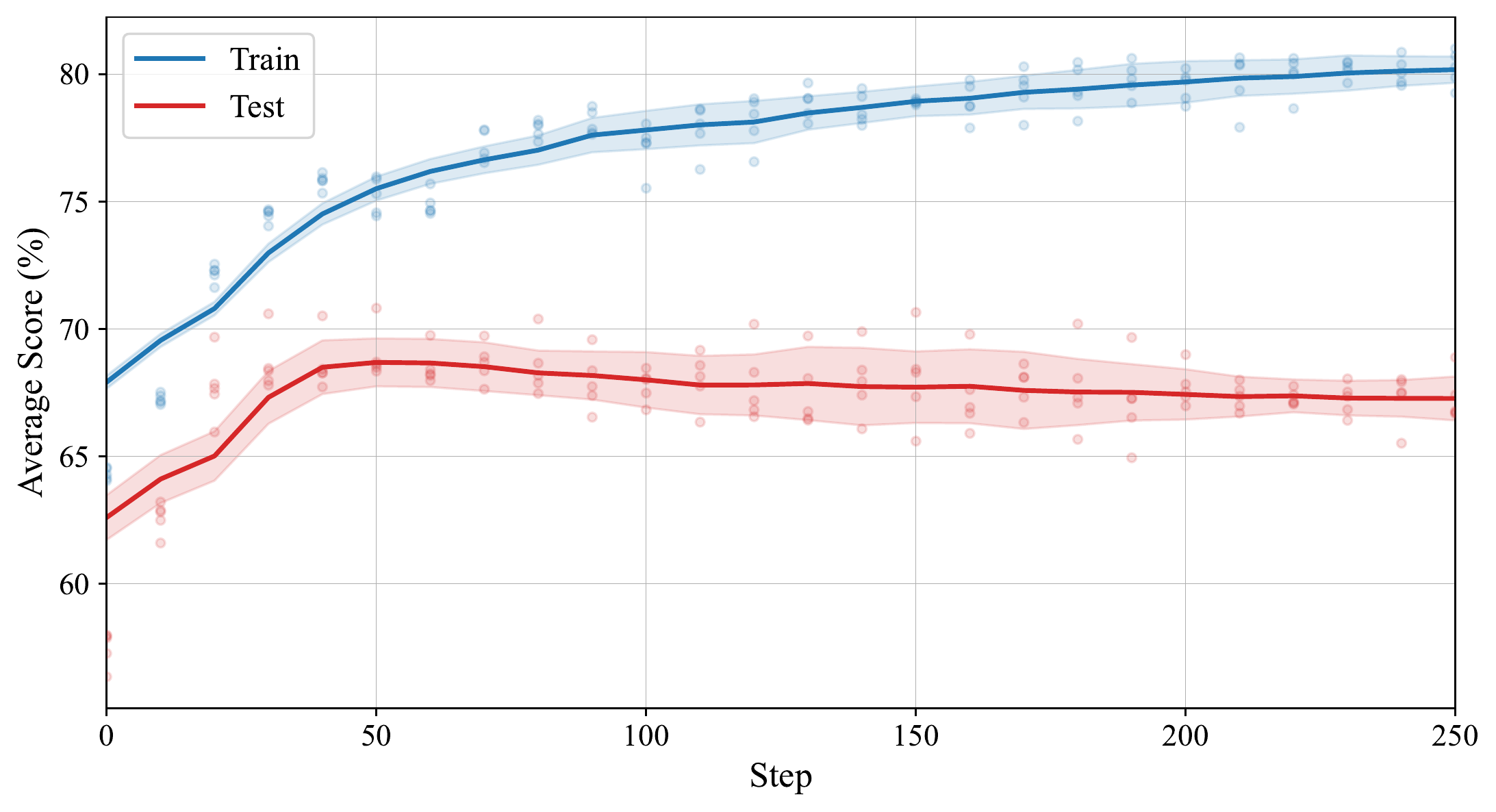}
    \caption{The training and test accuracy curves on EmbedLLM.}
    \label{fig:embedllm_acc}
\end{figure}

\paragraph{EmbedLLM}
We employ the official implementation~\citep{zhuang2024embedllm} and generate query embeddings using gte-qwen2-7B-instruct to ensure a fair comparison with our method, adjusting input layer dimensions accordingly. We increase the batch size to 32,768 for more stable training, keeping all other hyperparameters unchanged. Our implementation comprises approximately 12 million trainable parameters. The training and test accuracy curves are presented in Figure \ref{fig:embedllm_acc}.

\paragraph{Mixture of Agents} 
We adopt the classic three-layer architecture from the official implementation~\citep{wang2024mixture}, incorporating all ten candidate models (i.e., each proposer layer includes all ten models, with an aggregator layer summarizing the information). We choose Qwen2.5-7B-Instruct as the aggregator due to its strong performance, long context window, and robust instruction-following capabilities. To ensure that the context length stays within model limits, we set a context window of 6,000 tokens for each proposer layer and 20,000 tokens for the aggregator layer.

\paragraph{Mixture of Agents (Oracle)}
For a more comprehensive comparison, we enhance the MoA algorithm by adjusting the selection of models in each proposer layer. Specifically, rather than using all ten candidate models at each layer (to avoid performance degradation due to excessively long context windows), we manually selecte the three best-performing models on the validation set for each specific task as proposers. Nevertheless, Qwen2.5-7B-Instruct remains the aggregator across all tasks. For instance, for the AIME task, we select DeepSeek-Distill-R1-Qwen-7B and Qwen2.5-7B-Instruct as proposers, while for the MELD task, we choose Qwen2.5-7B-Coder-Instruct, Qwen2.5-7B-Instruct, and Fin-R1. The three-layer architecture and context window sizes remained consistent with the original MoA configuration.

\subsection{Symbolic Mixture-of-Experts}
We employ the official implementation~\citep{chen2025symbolicmixtureofexpertsadaptiveskillbased}, likewise utilizing Qwen2.5-7B-Instruct to generate skill keywords for each model and query. We adopt bge-m3 to generate embeddings for each keyword, thereby achieving superior performance compared to the Sentence-BERT used in the official implementation, while keeping all other hyperparameters unchanged.

\section{Implementation Details}
\label{sec:details}
All 22 LLMs are served on NVIDIA A800-80G GPUs. To speed up inference, every model—including the embedding model—is deployed with vLLM 0.8.4 on the same hardware. Direct (CoT) set $\text{temperature}=0.2, \text{top}\_p = 1.0$; voting strategies (Model-Switch, Self-Consistency) set $\text{temperature}=0.7, \text{top}\_p = 1.0$ (all other parameters remain at their default values). When using the aggregation strategy, we adopt the same parameters as in Direct (CoT).
For reproduce our results, we open-source our code on GitHub: \url{https://anonymous.4open.science/r/Avengers-2C0D}



\subsection{Benchmark Dataset Descriptions}
\begin{description}
  \item[AIME] Olympiad-level math benchmark of \emph{American Invitational Mathematics Examination} problems (2024 \& 2025); numeric-answer tasks covering algebra, geometry, number theory and combinatorics that push high-school contest reasoning.
  \item[MATH500] Subset of Hendrycks MATH: 500 diverse competition problems spanning probability, algebra, geometry and more, sampled for quick yet rigorous evaluation of multi-step symbolic reasoning.
  \item[LiveMathBench] 2024 benchmark with 140 recent questions from CNMO, CCEE, AMC and Putnam competitions, curated to avoid training contamination and probe stable mathematical reasoning in LLMs.
  \item[MBPP] Mostly Basic Python Programming: $\approx$1 000 crowd-sourced entry-level Python tasks, each with docstring, reference solution and three unit tests; used to measure code-generation functional correctness.
  \item[HumanEval] OpenAI HumanEval: 164 handwritten Python programming problems with hidden unit tests, specifically drafted to eliminate training leakage and evaluated via pass@k.
  \item[KORBench] Knowledge-Orthogonal Reasoning Benchmark evaluates intrinsic reasoning across five rule-driven categories (Operation, Logic, Cipher, Puzzle, Counterfactual) while minimizing domain-knowledge cues.
  \item[Knight and Knaves] Logic-puzzle dataset containing 700 scenarios with 2–8 inhabitants where knights always tell the truth and knaves always lie; the task is to deduce each role from their statements.
  \item[BBH] Big-Bench Hard: 23 of the most challenging BIG-Bench tasks where earlier models lagged human baselines, covering diverse domains such as causal reasoning, word sorting and formal fallacies.
  \item[ARCC] ARC Challenge—1172 multiple-choice grade-school science questions that simple retrieval baselines miss; assesses open-domain reasoning with a 14 M-sentence supporting corpus
  \item[MMLUPro] MMLU-Pro extends MMLU with 1000 harder questions across 14 disciplines plus chain-of-thought annotations, offering a more robust university-level multi-task understanding benchmark.
  \item[GPQA] Graduate-level Physics Qualifying Q\&A benchmark designed to be Google-proof; covers classical mechanics, EM, quantum, thermo and more, forcing deep conceptual reasoning.
  \item[FinQA] Financial numerical-reasoning dataset with 1147 Q\&A pairs over 2.8k corporate reports, combining tabular and textual evidence and requiring multi-step arithmetic reasoning.
  \item[MedQA] Medical QA benchmark built from USMLE and other board exams; 1273 English multiple-choice questions (plus Chinese variants) spanning basic science, diagnosis and treatment.
  \item[EmoryNLP] Emotion-in-conversation corpus from 97 Friends TV-show episodes, totalling 12 606 utterances labelled with seven discrete emotions for dialogue emotion recognition research.
  \item[MELD] Multimodal EmotionLines Dataset: 1.4 k multi-speaker dialogues and 13 k utterances from Friends, providing synchronized text, audio and vision plus emotion and sentiment labels.
\end{description}

\subsection{Model Descriptions}

\begin{description}
\item[Fin-R1] Specialises in quantitative reading of corporate data and long-form fiscal reasoning, while maintaining competitive scores on competition-style mathematics and factual QA. Its attention to ledger-like details comes at the cost of brittle performance on paradox-style logic games, underscoring a finance-oriented optimisation.
\item[Qwen2.5-7B-Instruct] Pairs top-tier coding accuracy with strong symbolic-math reasoning and achieves the most even spread across general-knowledge and dialogue tasks among its size class. Its moderate showing on intricate logic puzzles hints at a balanced—but not specialised—architecture that trades depth in any single niche for consistently high all-round competence.
\item[Qwen2.5-Coder-7B-Instruct] Excels at structured programming challenges and keeps pace with larger peers on algorithmic reasoning, while retaining solid performance in numeric problem-solving and business analytics. It concedes ground on tricky logical riddles and specialised domain questions, revealing a profile tuned first for code reliability rather than broad expert knowledge.
\item[DeepSeek-R1-Distill-Qwen-7B] Dominates Olympiad-level mathematics and logic riddles, while retaining competitive breadth across finance and encyclopaedic knowledge tests—a rare feat for its parameter count. The main gap is in highly specialised clinical material and affective-dialogue sensing, pointing to a math-centric curriculum.
\item[Llama-3.1-8B-Instruct] Balances solid software-engineering skills with reliable performance in factual and medical queries, and even handles tricky reasoning suites respectably. The trade-off is weaker abstract-math fluency and lower success on self-contradictory dialogue puzzles, placing it as a versatile generalist with coding leanings.
\item[Granite-3.1-8B-Instruct] Shows surprising strength on science-fact retrieval and spreadsheet-style arithmetic despite muted results on code and higher maths. The model is clearly tuned for concise knowledge lookup and tabular reasoning, but its conservative logical toolkit becomes evident on riddle-like tasks.
\item[Llama-3.1-8B-UltraMedical] Unrivalled within the cohort on clinical examination questions and therapeutic reasoning, confirming a medical specialist orientation. Outside its home turf it posts fair results on general reading but lags behind peers on advanced logic and algorithmic coding, illustrating the classic depth-versus-breadth trade-off.
\item[cogito-v1-preview-llama-8B] Stands out on open-ended reasoning and multi-domain science questions, matching larger models on financial and medical queries despite modest maths prowess. Slightly weaker numerical intuition is offset by a marked advantage on complex instruction-following and argumentative tasks, making it a ``thinker'' rather than a ``calculator''.
\item[gemma-2-9b-it] Offers dependable code generation and strong comprehension of technical prose, coupled with above-average domain knowledge in healthcare and finance. Elementary competition maths and deception-based puzzles remain its softer spots, signalling a language-focused training bias over abstract symbol manipulation.
\item[glm-4-9b-chat] Delivers well-rounded scores, peaking on encyclopaedic comprehension and practical coding, yet rarely leading except in fact-heavy domains. Its numbers reveal a steady, middle-of-the-pack stance—above smaller baselines in most categories, but outshone by maths specialists and domain experts on their respective turf.

\end{description}

\paragraph{Parameter Setting of Cluster Methods}
\label{sec:cluster_detailed}

We implement all clustering algorithms using \texttt{scikit-learn} with the following default hyperparameters:
\begin{itemize}[leftmargin=*]
\item \textbf{K-Means}~\citep{lloyd1982least, macqueen1967some}: We set the number of clusters $K=64$, with the ``k-means++'' initialization method, automatic initialization count (\texttt{n\_init=“auto”}), a maximum of 1000 iterations, and the “elkan” algorithm. 
\item \textbf{Hierarchical Clustering}~\citep{ward1963hierarchical} (Hier-Clustering): We use the Euclidean distance metric with the "ward" linkage method and set the number of clusters to $K=64$.
\item \textbf{Gaussian Mixture Models}~\citep{dempster1977maximum} (GMM): We specify 64 components, use full covariance matrices, and set the number of initializations (\texttt{n\_init}) to 10. The maximum iterations are set to 100, with verbosity level 2, printing intermediate results every 10 iterations. 
\item \textbf{Balanced Iterative Reducing and Clustering using Hierarchies}~\citep{zhang1996birch} (BIRCH): We use the default threshold of 0.5, a branching factor of 50, and set the number of clusters to $K=64$.
\item \textbf{Spectral Clustering} (Spec-Clustering): We set the number of clusters to $K=64$, use the RBF affinity matrix with the parameter gamma set to 0.1.
\end{itemize}

\subsection{Further Analysis}

\begin{figure}[!htbp]
    \centering
    \includegraphics[width=1\linewidth]{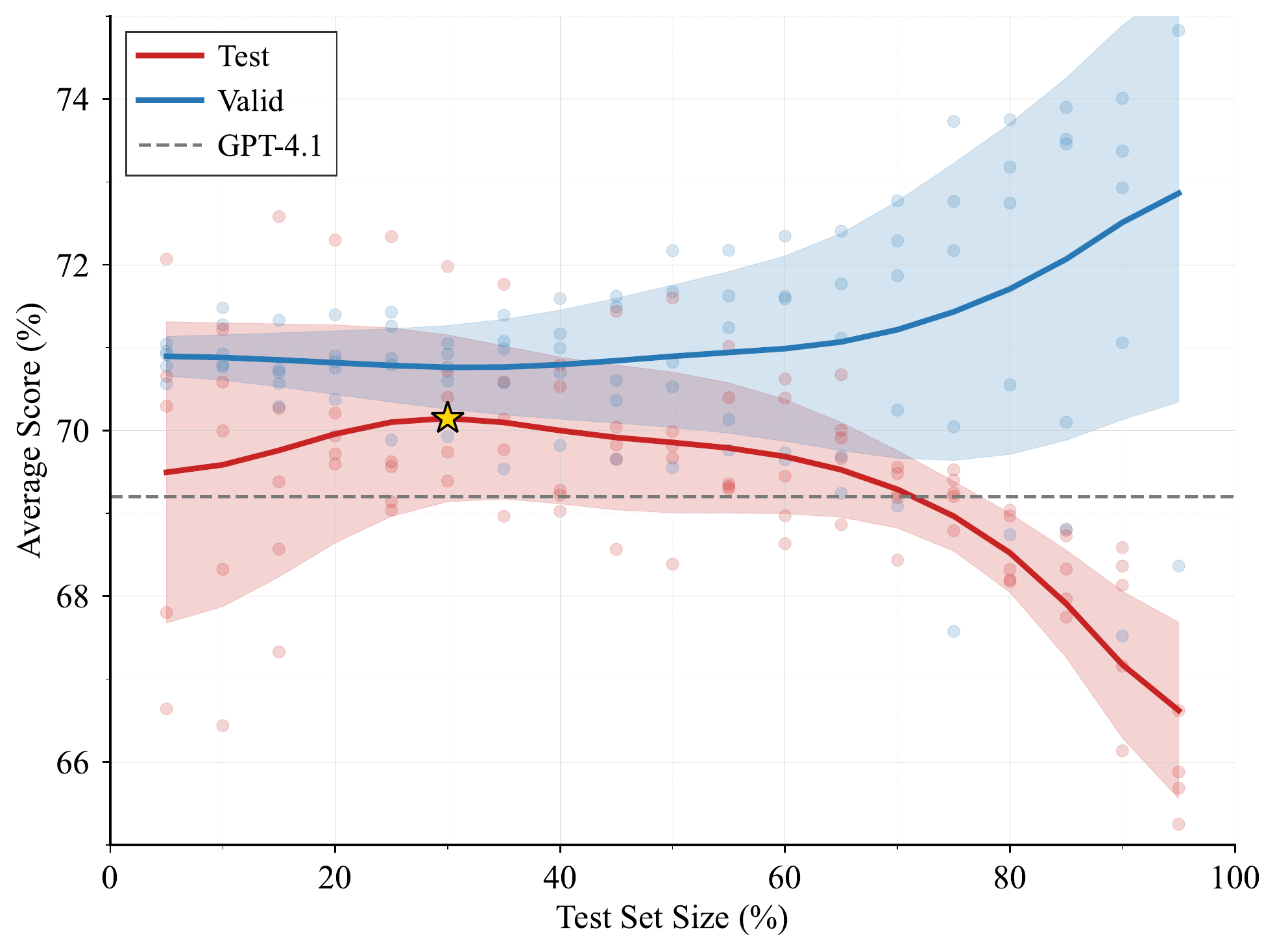}
    \caption{The impact of test set proportion on the \emph{Avengers} performance.}
    \label{fig:testsize}
\end{figure}

\paragraph{Test Size}
We further investigate how the proportion of the test set impacts the performance of the \emph{Avengers}. Specifically, we vary the test set proportion from 0.05 to 0.95 and analyze its effect on model performance. As shown in Figure~\ref{fig:testsize}, results indicate that the \emph{Avengers} achieves the best overall performance when the test set proportion is approximately between 0.3 and 0.35. Additionally, it is noteworthy that with only 30\% of the data used to fit the clustering model, the \emph{Avengers} already surpasses the performance of GPT-4.1, demonstrating the high efficiency of our method in terms of labeled data utilization.

\paragraph{Further Analysis of Ensemble Strategy}
In this paragraph, we further analysis the impact of ensemble strategy. As shown in Table~\ref{tab:ensemble-ablation-continue}, model-Switch achieves competitive overall performance, notably excelling on knowledge tasks by combining predictions from two models for complementary strengths. Among these, Self-Consistency emerges as the best strategy due to its simple yet effective procedure of majority voting, reducing single-sample inference errors and maintaining robustness across tasks.

\begin{table}[!h]
    \small
    \centering
        \caption{Ablation on the \emph{Avengers} ensemble strategies.  
    Numbers are the average accuracy for four categories (Mathematics, Logic, Knowledge, Affective); the Code category is omitted because there is no straightforward way to compute a self-consistency score for program generation.}
    \begin{tabular}{lccccc}
    \toprule
    \textbf{Ensemble Strategy} & \textbf{Mathematics} & \textbf{Logic} & \textbf{Knowledge} & \textbf{Affective} & \textbf{Avg.} \\
    \midrule
    Direct (CoT)  & 66.89 & 60.77 & 65.84 & 43.27 & 59.19 \\
    Aggregation & 59.39 & 53.89 & 68.91 & 44.75 & 56.73 \\
    Model-Switch   & \underline{71.11} & \underline{71.27} & \textbf{72.22} & \textbf{51.09} & \underline{66.42} \\
    Self-Consistency   & \textbf{73.42} & \textbf{72.06} & \underline{71.96} & \underline{50.50} & \textbf{66.98} \\
    \bottomrule
    \end{tabular}
    \label{tab:ensemble-ablation-continue}
\end{table}

\subsection{Speculative Insights on the Effectiveness of Clustering-Based Routing}
\label{app:insight}
In this paragraph, we would like to answer why our upscaling is effective without requiring training.
Overall, Figure~\ref{fig:k_impact} illustrates three distinct regimes:
\textbf{(i) Under-clustering} ($K < 14$): As the cluster number $K$ initially increases from very small values, both training and testing performance rapidly improve.
\textbf{(ii) Stability} ($14 < K < 140$): As $K$ grows further to a suitable intermediate range, both validation and test performance stabilize and remain closely matched. During this phase, the validation set performance improves gradually while the test set performance remains stable and consistently surpasses GPT-4.1. Notably, the cluster number ($K=64$) chosen for all our main experiments lies in the middle of this region, validating the robustness and stability of our parameter selection.
\textbf{(iii) Over-clustering} ($K > 140$): With a further increase in cluster number, the validation set performance gradually approaches the Oracle level, as each query nearly forms its own cluster, thus precisely finding the optimal model for each query. However, the performance on the test set sharply deteriorates, exhibiting typical over-clustering behavior.

A clear understanding of this “sweet-spot vs. over-clustering” pattern also offers a lens for interpreting why heavily trained router-based methods tend to struggle in our setting.
We hypothesize that other router-based methods (RouterDC, EmbedLLM and MODEL-SAT) requiring additional training are prone to quickly entering the over-clustering (overfitting) region illustrated in Figure~\ref{fig:k_impact}, due to the introduction of extra learnable parameters and complex network structures (such as embedding models, MLP layers, or even LLMs). These methods typically optimize directly toward maximizing validation-set (training) performance, and combined with their large number of parameters, this significantly accelerates overfitting and consequently harms their generalization performance. Moreover, identifying a simple yet effective early-stopping criterion to prevent overfitting becomes challenging in practice. This explains why, in Tables~\ref{tab:main-result} and~\ref{tab:ood-result}, we report peak performance on the test set rather than the validation set for these trained methods. In the Appendix~\ref{baseline}, we further provide training curves for these router-based methods, which exhibit a similar overfitting pattern to that observed in Figure~\ref{fig:k_impact}, further validating our hypothesis.

This experiment thus highlights the efficiency and simplicity of the \emph{Avengers}, which effectively balances the precision of model selection and generalization capability through clustering without introducing any learnable parameters. This design avoids the overfitting pitfalls common to more complex methods.

\subsection{Model Usage Distribution within the \emph{Avengers}}
\label{app:model-usage}

\begin{figure}
    \centering
    \includegraphics[width=1\linewidth]{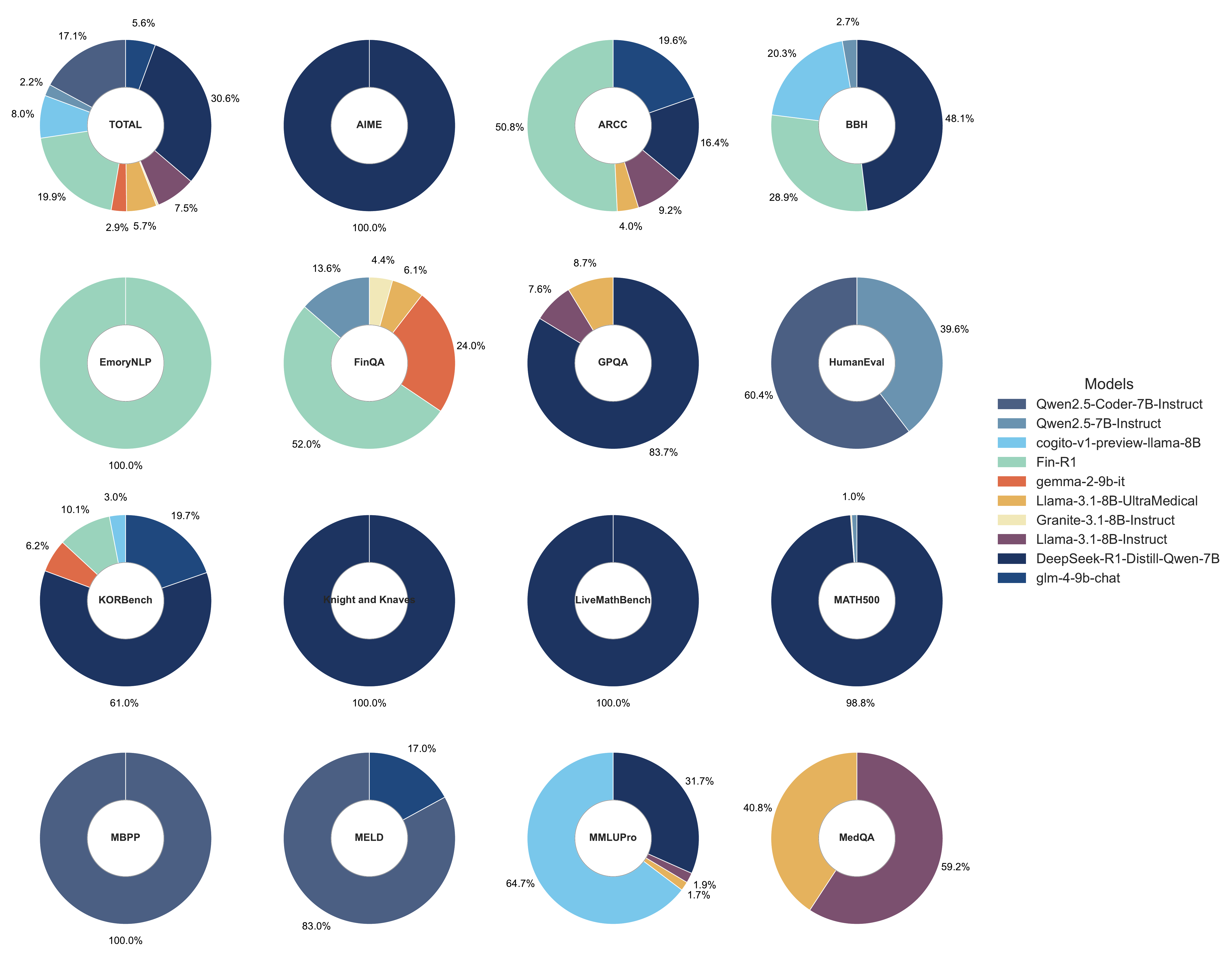}
    \caption{Distribution of the routing results of the \emph{Avengers} across tasks.}
    \label{fig:model-usage-distribution}
\end{figure}

Figure~\ref{fig:model-usage-distribution} provides a detailed breakdown of the model usage distribution within the \emph{Avengers} framework across various benchmark datasets. The analysis reveals that the \emph{Avengers} demonstrates a clear capability for automatically identifying models that balance strong performance with complementary strengths. While some well-known general-purpose models, such as Qwen2.5-7B-Instruct, dominate usage across multiple tasks, less prominent models also play critical roles for specific tasks. For instance, Fin-R1—originally developed for financial-domain tasks—is heavily utilized in the ARC Challenge ($\sim $50\% of queries), while LLaMA-3.1-8B-UltraMedical, developed by a small academic team, is notably selected for approximately 40\% of queries in the MedQA dataset. These results underscore not only the versatility and effectiveness of the \emph{Avengers} approach but also highlight the significant yet often overlooked potential of specialized, community-contributed open-source models, contributing towards a more diverse and inclusive AI ecosystem.

\subsection{Limitation}
\label{app:limitation}
Similar to other router-based methods, the \emph{Avengers} requires setting aside a portion of the data as a validation set to obtain capability profiles.

Although we conduct experiments across 15 datasets spanning five task categories and performed additional out-of-distribution evaluations on five extra datasets, the \emph{Avengers} has not yet been validated on chat-based tasks involving cross-cultural, multilingual, or conversational contexts. This limitation arises primarily due to the subjective nature of these tasks, which lack standardized objective metrics, making their performance challenging to measure reliably.

Currently, our experiments with the \emph{Avengers} focus exclusively on integrating the capabilities of multiple small-scale (approximately 7B) language models, and we have not yet explored its applicability to smaller-scale models. The choice of 7B-scale models is primarily motivated by the abundant availability and active community support for open-source models of this scale.



\end{document}